\pdfoutput=1
\documentclass[11pt]{article}

\usepackage[preprint]{acl}

\usepackage{times}
\usepackage{latexsym}

\usepackage[T1]{fontenc}
\usepackage[utf8]{inputenc}
\usepackage{microtype}
\usepackage{inconsolata}

\usepackage{graphicx}
\usepackage{amsmath}
\usepackage{amssymb}
\usepackage{booktabs}
\usepackage{algorithm}
\usepackage{algpseudocode} 
\usepackage{array, multirow}
\usepackage{xcolor}
\usepackage{amssymb}
\usepackage[percent]{overpic}
\usepackage{enumitem}
\usepackage{CJKutf8}
\usepackage{xcolor}
\usepackage{svg}
\usepackage{subfig}
\usepackage{url}
\usepackage{comment}
\usepackage{makecell}

\usepackage[capitalize]{cleveref}

\crefname{section}{Sec.}{Secs.}
\Crefname{section}{Section}{Sections}
\Crefname{table}{Table}{Tables}
\crefname{table}{Tab.}{Tabs.}

\usepackage{xspace} % To handle spacing after macros smartly

\setlist[enumerate]{
  label={\upshape(\roman*)},
  labelwidth=*,
  nosep
}

\newcommand{\paratitle}[1]{\smallskip\noindent\textbf{#1}}
\newcommand{\eg}{\textit{e.g.,} \xspace}

\newcommand{\ie}{\textit{i.e.,}\xspace}

\title{Evaluating LLM Adaptation to Sociodemographic Factors: \\ User Profile vs. Dialogue History}

\author{
  Qishuai Zhong\textsuperscript{1} \quad
  Zongmin Li\textsuperscript{1} \quad
  Siqi Fan\textsuperscript{2} \quad
  Aixin Sun\textsuperscript{1} \\
  \textsuperscript{1} Nanyang Technological University, Singapore \\
  \textsuperscript{2}University of Electronic Science and Technology of China, Chengdu, China
}

\begin{document}
\maketitle
%\maketitle

%%%%%%%%% ABSTRACT
\begin{abstract}
Effective engagement by large language models (LLMs) requires adapting responses to users’ sociodemographic characteristics, such as age, occupation, and education level. While many real-world applications leverage dialogue history for contextualization, existing evaluations of LLMs’ behavioral adaptation often focus on single-turn prompts. In this paper, we propose a framework to evaluate LLM adaptation when attributes are introduced either (1) explicitly via user profiles in the prompt or (2) implicitly through multi-turn dialogue history. We assess the consistency of model behavior across these modalities. Using a multi-agent pipeline, we construct a synthetic dataset pairing dialogue histories with distinct user profiles and employ questions from the Value Survey Module (VSM 2013)~\cite{vsm2013-jj} to probe value expression. Our findings indicate that most models adjust their expressed values in response to demographic changes, particularly in age and education level, but consistency varies. Models with stronger reasoning capabilities demonstrate greater alignment, indicating the importance of reasoning in robust sociodemographic adaptation.

\end{abstract}

\section{Introduction}
\label{sec:introduction}

Large Language Models (LLMs) and their chatbot applications have garnered significant attention since the release of ChatGPT~\cite{openai2024gpt4technicalreport, dam2024completesurveyllmbasedai}. Numerous open-source frameworks~\cite{wolf2020huggingfaces, vllm} now enable organizations to deploy LLM on-premises across various domains. For AI service providers, hosted models are expected to align responses appropriately with users’ sociodemographic characteristics, as well as their cultural and ethical preferences, to enhance engagement and interaction quality~\cite{sicilia2024humbelhumanintheloopapproachevaluating, altenburger2024examiningrolerelationshipalignment}.

Unlike humans, who tend to maintain consistent values, ethics, and social norms across contexts, studies have shown that LLMs exhibit variability in their expressed values, which are learned from human-generated training data and shaped by contextual cues~\cite{kharchenko2024llmsrepresentvaluescultures, kovač2023large}. This variability presents a risk: LLMs may inadvertently perpetuate harmful stereotypes, such as labeling Generation Z as “Digital Addicts”~\cite{twenge2017smartphones}. To mitigate such issues and foster user trust, LLMs should dynamically tailor their responses to reflect user expectations—a capability we refer to as \textbf{behavioral adaptation}.

\begin{figure}
\centering
%\fbox{
    \includegraphics[trim=1cm 6.5cm 18.5cm 0.5cm,clip=true,width=0.475\textwidth]{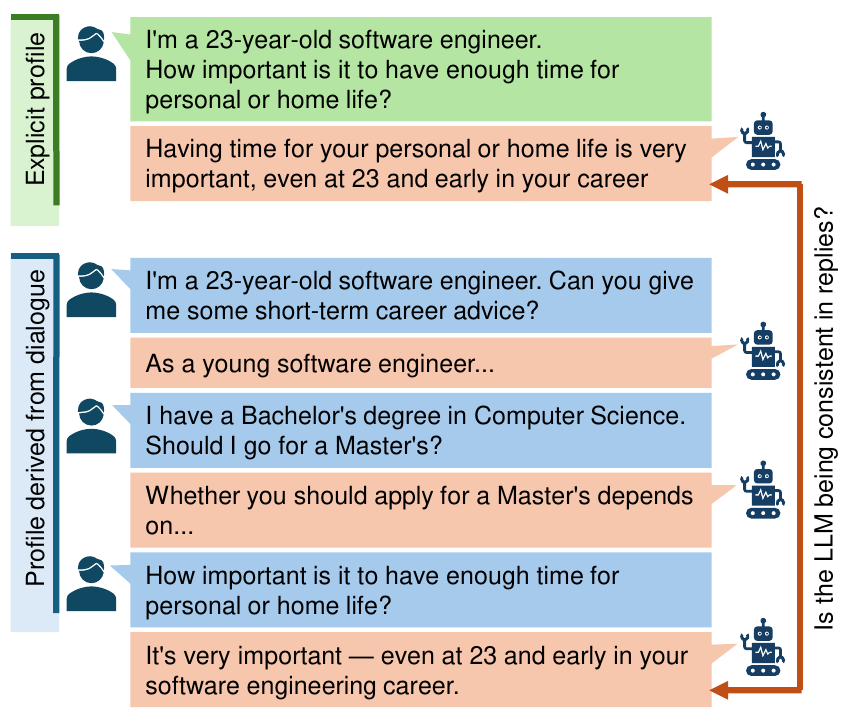}
   % }
    \caption{We evaluate whether the model can adjust response values according to identical user attributes presented in different formats, and assess the consistency across these formats.}
\label{fig:research_target}
\end{figure}

Sociodemographic attributes of user profiles (e.g., age, education, occupation, nationality) are strongly correlated with cultural norms and values related to family, authority, and social behavior~\cite{Fung2016-re, genderandvalues, Gelfand2011-ei}. Recent work has explored value alignment between LLMs and user profiles~\cite{yao2024claveadaptiveframeworkevaluating, zhang2024heterogeneousvaluealignmentevaluation, sukiennik2025evaluationculturalvaluealignment}. However, these studies largely focus on single-turn inputs where user profiles are explicitly provided in the prompt (see Figure~\ref{fig:research_target}, top). This leaves a gap in understanding whether LLMs maintain behavioral consistency when profiles are instead inferred implicitly through dialogue (see Figure~\ref{fig:research_target}, bottom).

We identify two key challenges in such dialogue-based evaluation:
(i) Can LLMs accurately infer demographic attributes from chat history?
(ii) If so, can they adapt their responses accordingly?

Dialogue history provides essential context for identifying user traits in real-world applications~\cite{dam2024completesurveyllmbasedai}. Prior work has developed datasets to evaluate this capability. In our study, we leverage the FaithfulPersonaChat benchmark~\cite{faithfulPersonaChat} to assess persona recognition in \texttt{Llama3.1-8B-Instruct} as a representative LLM (see Appendix~\ref{appendix:persona_recognition}). Our findings confirm that the model can reliably infer at least one persona attribute from multi-turn dialogue, partially addressing challenge (i).

Building on this, we propose a novel evaluation framework to quantify how LLMs adapt their value expression when presented with demographic information either explicitly (via user profile) or implicitly (via dialogue), to address challenge (ii). However, existing benchmarks lack dialogues annotated with demographic attributes, which are crucial for controlled comparisons. To address this, we introduce an agent-based generation pipeline that constructs an evaluation dataset with aligned sociodemographic attributes across both input formats. In summary, our contributions are threefold: 
\begin{itemize}
    \item We introduce an evaluation framework that assesses LLM value adaptation across two input formats, (i) single-turn prompts with explicit user profiles, and (ii) multi-turn dialogues where profiles are embedded implicitly, and measures consistency across both.
    \item We present a novel, agent-based dataset construction method for generating profile-aligned dialogue data.
    \item We evaluate multiple open-source LLMs using the Value Survey Module (VSM 2013)~\cite{vsm2013-jj} to measure value expression.
\end{itemize}

Our experiments show that most models adjust their expressed values in response to demographic changes, especially in age and education level. Moreover, the degree of value adjustment is positively correlated with the magnitude of attribute change. However, consistency across input formats varies by model. Smaller models exhibit greater variability, while larger models with stronger reasoning capabilities show better alignment across formats. Notably, reasoning-augmented models like \textit{QwQ-32B}~\cite{qwen2025qwen25technicalreport} achieve the highest consistency, underscoring the critical role of reasoning in robust sociodemographic adaptation.

%====================
\section{Literature Review}
%====================
This study examines LLM behavior adaptation in multi-turn human-model interactions and assesses value consistency across profile presented conditions. Given its intersection with persona attribute extraction and cultural value alignment in LLMs, we survey related work in both fields.

\subsection{Persona Attributes Understanding}

Evaluations of language models’ understanding of persona attributes typically center on two tasks: next‐utterance prediction and persona expansion. Standard benchmarks such as PersonaChat~\cite{personachat}, RealPersonaChat~\cite{realpersonachat}, and FaithfulPersonaChat~\cite{faithfulPersonaChat} provide dialogues annotated with descriptive persona statements (\eg “I’m a pet lover”) for these tasks. Other efforts, like Pchatbot~\cite{pchatbot}, compile large‐scale Chinese dialogues from Weibo and judicial forums but lack explicit demographic mappings. LiveChat~\cite{gao-etal-2023-livechat} augments live‐stream conversations with streamer personas that include demographic attributes, yet this information serves only as auxiliary context for next‐utterance prediction.

Despite these resources, most datasets consist of human–human dialogues embedding persona descriptions rather than demographic profiles, hindering controlled analysis of LLM adaptation to quantifiable attributes. For example, grouping by age is straightforward, but contrasting more abstract traits—such as “running enthusiast” versus ``someone who lost a dog''—yields unreliable comparisons. To overcome this limitation, we construct a synthetic dataset specifically tailored for rigorous evaluation of sociodemographic adaptation.

%=====================================
\subsection{Evaluating Values of Models } 
%=====================================

Several studies have evaluated LLMs on how they express social and cultural values in response to different prompts. A common approach involves using research instruments like Hofstede's Value Survey Module (VSM)~\cite{vsm2013-jj}, which has been applied in prior work~\cite{kharchenko2024llmsrepresentvaluescultures, arora2023probing, masoud2024culturalalignmentlargelanguage} to assess whether models align their responses with cultural contexts. Despite differences in methodology, findings consistently show that LLMs adjust their value expressions based on contextual cues.

Other studies have constructed evaluation datasets based on the World Values Survey~\cite{Haerpfer2020-qe}, including GlobalOpinionQA~\cite{durmus2024towards} and WorldValueBench~\cite{zhao2024worldvaluesbenchlargescalebenchmarkdataset}. The former also incorporates value questions from Pew\footnote{\url{https://www.pewresearch.org/}} finding that most LLMs tend to favor Western perspectives. The latter focuses on evaluating models' awareness of demographic contexts, revealing that even advanced LLMs struggle to capture the nuances of multicultural value systems.

Recent studies have investigated model values within specific contexts. BiasLens~\cite{li2024benchmarkingbiaslargelanguage} systematically examines social biases in LLMs through role-playing scenarios, while \citet{moore-etal-2024-large} evaluates value consistency across prompt variations, revealing generally stable model outputs. \citet{liu2025llmsgraspimplicitcultural} introduces a benchmark dataset to assess LLMs' ability to infer implicit cultural values from natural conversational contexts, emphasizing the challenges of nuanced attitude detection and open-ended cultural reasoning.

Unlike prior studies, we analyze LLMs' value expression patterns using multi-turn human–model interactions, which better reflect real-world chatbot inputs. We also evaluate behavioral consistency when sociodemographic attributes are provided explicitly versus implicitly within the dialogue.

%==============================
\section{Research Targets}
\label{sec:research_targets}
%==============================

We define \emph{behavior adaptation capability} as a model’s ability to adjust response values and tone in accordance with users' sociodemographic attributes. A key focus of our study is to examine whether this adaptation remains consistent when the same demographic information is supplied explicitly in a single‐turn prompt versus implicitly via earlier dialogue (see Figure~\ref{fig:research_target}).

We use the Value Survey Module (VSM 2013)~\cite{vsm2013-jj}, grounded in Hofstede's Cultural Dimensions Theory~\cite{Gerlach2021-ww}, to quantify cultural values. This questionnaire features multiple-choice items on workplace dynamics and decision-making, each with five options (IDs 1–5). From the original 24 items, we select an 18-question subset \(Q\), omitting emotional and health-related items. We apply this survey to evaluate model behavior adaptation across below experimental scenarios.

\paratitle{Behavior Adaptation to User Profile (\texttt{BA\_user})}: This scenario evaluates whether models can adjust their responses based on explicit user profile consisting of sociodemographic attributes presented in the context, \eg ``Answer questions based on the given user profile: age: 23, job title: data scientist, gender: male, education: Bachelor's degree.''
   
\paratitle{Behavior Adaptation to Dialogue History (\texttt{BA\_dialogue})}: Instead of relying on explicit attributes, models are tested on their ability to infer and adapt from dialogue history~\cite{gupta-etal-2024-llm}. This scenario mimics real-world interactions, where the model must interpret user intent and context from prior exchanges.

\paratitle{Consistency Across Profile and Dialogue History (\texttt{Consistency})}: Beyond behavioral adaptation, our framework also evaluates whether models maintain behavioral consistency when processing equivalent user attributes presented in different representational formats. Specifically, we expect models to respond similarly to the value survey when the same demographic attributes are provided through explicit user profiles or implicitly inferred from dialogue history.

%============================
\section{Dialogue Dataset Generation}
%============================
\label{sec:dataset_generation}

\begin{figure}
\centering
%\fbox{
    \includegraphics[trim=0.5cm 11.5cm 18.5cm 0.5cm,clip=true, width=0.4\textwidth]{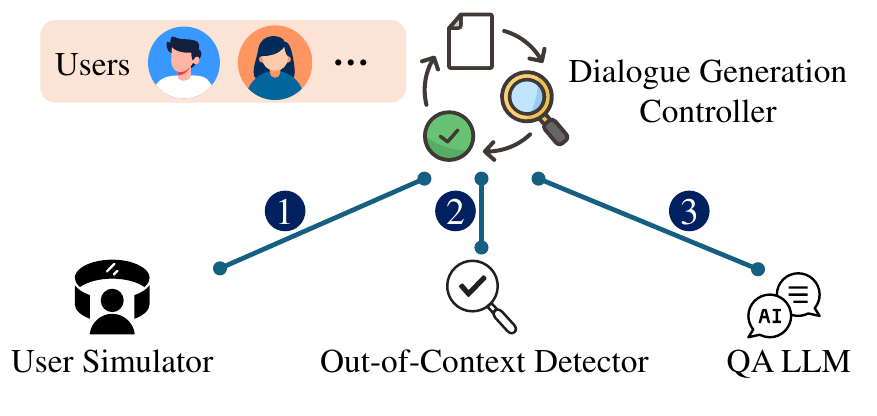}
%}

    \caption{Dataset generation framework architecture. Each iteration: (i) user\_simulator LLM is queried to generate a question simulating the user's perspective based on their profile, (ii) out-of-context detector validates the question to ensure consistency with the user's profile, and (iii) qa\_llm responds to the question.}
\label{fig:conversation_generation}
\end{figure}

While \texttt{BA\_user} assessment is straightforward, evaluating \texttt{BA\_dialogue} and \texttt{Consistency} is hindered by the absence of datasets meeting two key criteria: (1) human–model dialogues with organically embedded demographic attributes, and (2) explicit mappings between each dialogue and its corresponding user profile.

Drawing inspiration from prior works~\cite{abdullin2024syntheticdialoguedatasetgeneration, chen-etal-2024-diahalu}, we design a multi‑agent workflow to generate synthetic, career‑advice dialogues from user profiles sourced from a curated simulated dataset.\footnote{\url{https://www.kaggle.com/datasets/ravindrasinghrana/employeedataset/data}} Each dialogue maps to a unique profile, embedding demographic attributes—age, education, occupation, and nationality—within contextually grounded interactions. This career‑focused domain aligns with our value assessment framework, enhancing interpretability in downstream evaluations.

The dialogue is generated iteratively by the workflow under the supervision of a generation controller, as illustrated in Figure~\ref{fig:conversation_generation}. The controller orchestrates three key LLM components:

\paratitle{User Simulator (\texttt{user\_simulator})} 
We employ \texttt{Gpt-4o-2024-08-06}~\cite{openai2024gpt4technicalreport} to emulate a user seeking career advice via question–answer (QA) interactions with an LLM. Each query generated by the simulator is guided by: (1) user demographic attributes for personalization, (2) instructions and preceding dialogue, guiding the generation of contextually relevant queries, and (3) predefined conversation objectives and termination criteria. The simulator ends the dialogue once the specified termination condition is satisfied.

\paratitle{Out-of-Context Detector (\texttt{ooc\_detector})}: We employ \texttt{Gpt-4o-mini-2024-07-18}~\cite{openai2024gpt4technicalreport} to validate the questions generated by the user simulator. It ensures that each question aligns with the user’s profile and maintains consistent first-person framing. If inconsistencies are detected, the \texttt{ooc\_detector} directly revises the question.

\paratitle{Question Answering LLM (\texttt{qa\_llm})}: The LLM responds to the simulated user's queries with its default configuration to ensure natural interactions. To replicate real-world chatbot behavior, past dialogue history is always prepended to the latest user question, following standard practices for context injection in human-LLM communication.

Each generation loop terminates when the user simulator decides to conclude the dialogue or the maximum iteration limit (\texttt{max\_runs}) is reached.

More details on the prompt design for each component are provided in Figure~\ref{fig:user_simulator_prompts} (Appendix~\ref{appendix:dataset_generation_details}), and the complete conversation generation procedure is outlined in Algorithm~\ref{algo:dialogues_generation_procedure} (Appendix~\ref{appendix:code}). In total, 1000 dialogue sets, denoted by \(D\), are generated,\footnote{\url{https://github.com/FerdinandZhong/model_behavior_adaption}} each mapped to a unique user from the seed dataset. The full user set is denoted by \(U\).

%========================================

\begin{table}
\centering
\small
\begin{tabular}{l|c|c}
\toprule
\textbf{Dimensions} & { \textbf{LLM Judge} } & {\textbf{Human}} \\
\midrule
{Attribute Coverage} & 4.14 & 3.64 \\
{Attribute Correctness} & 4.76 & 4.97 \\
{Question Diversity} & 4.52 & 4.65 \\
{Relevance} & 4.63 & 4.26 \\

\bottomrule
\end{tabular}
\caption{Overall ratings for generated dialogues by the LLM judge and 50 human‐rated samples. Despite stricter human judgments, high scores across all four dimensions confirm dialogue quality.}

\label{tab:rating}
\end{table}

\subsection{Dataset Evaluation}  
To assess the quality of the generated dataset, we conduct both human and LLM-based evaluations. LLM evaluation follows the widely adopted ``LLM-as-a-judge'' methodology~\cite{judgingllmasajudge, gu2025surveyllmasajudge}. Only questions generated by the \texttt{user\_simulator} are assessed, as the \texttt{qa\_llm} role-plays as itself to respond to these simulated queries. Assessments are conducted across four dimensions:

\paratitle{Attribute Coverage.} The number of demographic attributes explicitly mentioned (up to 5).

\paratitle{Attribute Correctness.} The number of correctly referenced demographic values. For example, if the user's age, gender, and job title are mentioned, but an incorrect age is used, the score for Attribute Coverage is 3 and for Attribute Correctness is 2.

\paratitle{Question Diversity.} The variety of topics covered, reflects the simulator's ability to generate distinct, contextually rich questions. For example, if four questions are generated but all focus solely on short-term career advice, then the score is 1.

\paratitle{Relevance.} The extent to which the questions remain contextually appropriate for career advice, and align with the \texttt{qa\_llm}'s prior responses to maintain coherent conversational flow.

For human evaluation, three postgraduate annotators score the generated questions on a 0 to 5 scale, with higher values indicating better quality. All annotators independently assess the same subset of 50 randomly selected samples. For automated evaluation, we use \texttt{gpt-4o-2024-08-06} as the judge to score all samples. Identical scoring guidelines are supplied to both human and LLM raters (Appendix~\ref{appendix:judge_prompt}), in accordance with best practices from~\citet{leng_llm_2023}. To reduce variability, the LLM judge evaluates each sample with 10 different random seeds, and we report the average score.

We assess alignment between average human and LLM judge ratings on the shared subset using the Pearson correlation coefficient~\cite{freedman2007statistics} and the two-way mixed-effects intraclass correlation coefficient (ICC(3,\emph{k}))~\cite{Shrout1979-ICC}. We omit Fleiss’ Kappa due to its sensitivity to category prevalence in our skewed data~\cite{Hoehler2000BiasAP}. Results (Appendix~\ref{appendix:alignment}) demonstrate strong concordance between human and automated evaluations. Summary statistics for all four evaluation dimensions are presented in Table~\ref{tab:rating}, confirming the high quality of the generated dataset.

%======================================
\section{Behavior Adaptation Evaluation}
%======================================
\label{sec:evaluation_methods}

Using our synthetic dataset, we assess LLM behavior adaptation via two scenarios.

In the \texttt{BA\_user} scenario, each time a model is queried with a VSM question \(q_j \in Q\) and an explicit user profile \(u \in U\), where \(q_j\) denotes the $j$th question. The model must return a \texttt{selected\_option\_id} (1–5) corresponding to the question’s \texttt{option\_ids}—the IDs of available choices—along with a justification and the log probability distribution over \texttt{option\_ids}. We denote this composite response by \(r_u^j\).

\begin{figure*}
\centering
%\fbox{
    \includegraphics[trim=0.8cm 12.5cm 16cm 1.2cm,clip=true,width=0.75\textwidth]{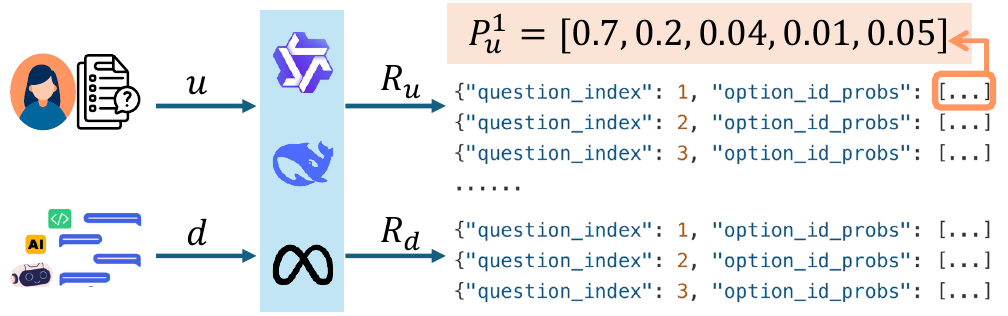}
 %   }
    \caption{Model querying workflow with key components. Here, \(u\) denotes a user profile, and \(d\) is a synthetic dialogue. 
    Each line \(r\in R\) represents the response to a VSM question, which includes a normalized probability distribution \(P\) over the 5 \texttt{option\_ids}.}
    
    \label{fig:querying_model_workflow}
\end{figure*}

In the \texttt{BA\_dialogue} scenario, each time a model is presented with a synthetic dialogue \(d \in D\) followed by a VSM question \(q_j\), it generates a response \(r_d^j\) under the same requirements. The complete sets of responses \(\{r_u^j\}\) and \(\{r_d^j\}\) for each model are denoted by \(R_U\) and \(R_D\), respectively. The querying workflow is illustrated in Figure~\ref{fig:querying_model_workflow}.

\subsection{Distance Definition}
\label{sec:distance_definition}

The core component of each response \(r\) is a normalized probability distribution \(P\) over all possible \texttt{option\_ids}. This distribution is computed from the model's log probability outputs (see Appendix~\ref{appendix:normalization} for details). For instance, if a response \(r\) has a selected \texttt{option\_id} of 2, the distribution might be \([0.1, 0.7, 0.05, 0.0, 0.15]\), where each value represents the relative likelihood of the corresponding \texttt{option\_id}. After normalization, the total probability sums up to 1.

We quantify the unit divergence between two responses \(r\)'s by calculating the Jensen-Shannon divergence (JSD) between their corresponding probability distribution \(P\)'s. That is, \(\text{JSD}(P_u^j \parallel P_{{u'}}^j)\) donates the divergence between \(r_u^j\) and \(r_{u'}^j\) where \(u\) and \(u'\) refer to two distinct user profiles.

To assess \texttt{BA\_user} and \texttt{BA\_dialogue}, both \(U\) and \(D\) are partitioned into groups \(g\) based on attributes including age, education, occupation, and nationality. 
% Group divergences are used to quantify the extent to which models adjust their responses when faced a different demographic profiles. 
For each group and question index \(j\), we compute the Jensen–Shannon centroid \(c^j\)—the distribution minimizing the total JSD to all group responses \(P_i^j\)~\cite{Jensen_Shannon_Centroid}. Starting from the mean distribution \(\bar{P}^j\), the centroid is obtained by
\[
c^j = \arg\min_{c}\,\sum_{i=1}^{n} \mathrm{JSD}\bigl(c \,\|\, P_i^j\bigr),
\]
where \(P_i^j\) is the \(i\)th response distribution for question \(j\) in a group of size \(n\).

The overall divergence between two groups \(g\) and \(g'\) is then defined as:
\[
    \text{Distance}(g, g') = \frac{1}{|J|} \sum_{j=1}^{|J|} \text{JSD}(c^j_g \parallel c^j_{g'}),
\]
This centroid-based approach helps mitigate the effects of outliers and uneven group sizes.

We establish baseline values for both \texttt{BA\_user} and \texttt{BA\_dialogue} using a consistent methodology. To illustrate, consider the baseline for \texttt{BA\_user} evaluation. After grouping \(R_U\)  by sociodemographic attribute and computing a centroid for each group, we define the global centroid as the Jensen–Shannon centroid of all responses in 
\(R_U\). The baseline is then calculated as the average distance between each group centroid and the global centroid. Comparing this baseline to the inter-group divergences allows us to quantify the model’s sensitivity to demographic variation. For example, when grouping \(R_U\)  by age, the divergence between the ``<30'' and ``>60'' cohorts should noticeably exceed the baseline, while the divergence between ``<30'' and ``30–40'' should remain below it—illustrating that greater age gaps drive greater variation in generated responses.

To evaluate the \texttt{Consistency} scenario, we compare responses \(R_u\) and \(R_d\) corresponding to the same user, by using their selected \texttt{option\_id} sequences rather than the full probability distributions. Let  
\(S_u = [s_1, s_2, \dots, s_m]\), and 
\(S_d = [t_1, t_2, \dots, t_m]
\),
where \(s_j\) is the selected \texttt{option\_id} in 1 to 5 for the  \(j\)th question (the same applies to \(t_j\)),  and \(m\) is the total number of questions, and \(m=18\) in our setting.

Since the \texttt{option\_ids} are ordinal and questions are independent, we employ the Earth Mover’s Distance (EMD)~\cite{EMD} to quantify alignment between \(S_u\) and \(S_d\). Let \(h_u(k)\) and \(h_d(k)\) denote the frequencies of value \(k\) in \(S_u\) and \(S_d\), respectively, and let \(K=5\) be the total number of discrete bins. Then
\[
\mathrm{EMD}(S_u, S_d)
= \sum_{x=1}^{K-1} \biggl\lvert \sum_{k=1}^{x}\frac{h_u(k)}{m}
  - \sum_{k=1}^{x}\frac{h_d(k)}{m} \biggr\rvert.
\]

We define a model’s \texttt{Consistency} as the mean EMD across all matched user profile $u$ and his/her dialogue $d$ pairs or \((u,d)\)'s, with a baseline computed over an equivalent number of random \((u,d)\) pairings \ie randomly matching a user profile with another user's dialogue. This metric enables direct comparison of consistency when the same profile is presented in different formats.

Table~\ref{tab:measure_forms} provides the details of the measured divergences and baseline values for all scenarios.

% +++++++++++++++++++++++++++++++++++++++++++++++++++++++++++++++
\begin{table*}[t]
\centering
\small
\begin{tabular}{l|l|l}
\toprule
    \textbf{Scenarios} &  \textbf{Measured Distance}  & \textbf{Baseline} \\
    \midrule
    \texttt{BA\_user}\  & 
    \(
    \frac{1}{|J|} \sum_{j=1}^{|J|}   \text{JSD}\left(c^j_g \parallel c^j_{g'}\right), \quad g, g' \subseteq U
    \)  &
    \(\frac{1}{|J|}\frac{1}{|G|} \sum_{j=1}^{|J|}\sum_{g=1}^{|G|}   \text{JSD}\left(c_g^j\parallel c^j_{U}\right)\)
    \\
    \midrule
    \texttt{BA\_dialogue}\  & 
    \(
    \frac{1}{|J|} \sum_{j=1}^{|J|}   \text{JSD}\left(c^j_g \parallel c^j_{g'}\right), \quad g, g' \subseteq D
    \)  &
    \(\frac{1}{|J|}\frac{1}{|G|} \sum_{j=1}^{|J|}\sum_{g=1}^{|G|}   \text{JSD}\left(c_g^j \parallel c^j_{D}\right)\)
    \\
    \midrule
    \texttt{Consistency}\  & 
    \( \frac{1}{|U|} \sum_{i=1}^{|U|}
      \text{EMD}\left(S_u, S_d \right) \)&
    % \(\frac{1}{2}\left(\frac{1}{|J|}\frac{1}{|D|} \sum_{j=1}^{|J|}\sum_{d \in D } JSD(P^j_{d} \parallel c^j_{D})\right)\)
    \(\frac{1}{|U|} \sum_{i=1}^{|U|}
    \mathbb{I}(u_i \not\to d_i)  \text{EMD}\left(S_u, S_d \right)\)
    \\

\bottomrule

\end{tabular}
\caption{List of quantified divergences and baselines across evaluation scenarios. For \(\texttt{BA\_user}\) and \(\texttt{BA\_dialogue}\), responses are partitioned into attribute-based groups \(g\) (\textbf{age}, \textbf{education level}, \textbf{occupation}, and \textbf{country}) to assess their influence on model behavior. Scenario \(\text{Consistency}\) measures the alignment between user profiles \(u\) and their corresponding dialogue histories \(d\). \(J\) represents the set of questions. \(u_i \not\to d_i\) means the user profile \(u_i\) and dialogue \(d_i\) are not from a same user.} 

\label{tab:measure_forms}

\end{table*}
% +++++++++++++++++++++++++++++++++++++++++++++++++++++++++++++++

%==========================
\subsection{Experiment Setting}
%==========================

Following the workflow outlined in Figure~\ref{fig:querying_model_workflow}, we evaluate multiple open-source LLMs, including the \textit{Qwen2.5} family, \textit{Llama3.1} family, \textit{DeepSeek-V3}, and the reasoning model \textit{QwQ-32B}~\cite{qwen2025qwen25technicalreport, dubey2024llama3herdmodels, deepseekai2024deepseekv3technicalreport}, to generate \(R_U\) and \(R_D\). Model outputs are structured in JSON format using XGrammar~\cite{dong2024xgrammarflexibleefficientstructured} for streamlined processing. Each model is queried once per prompt, except for the reasoning model \textit{QwQ-32B}, which is queried twice. The first query is unconstrained to allow for reasoning content generation, which is then appended to the second query to elicit a structured response. Full prompt designs are listed in Appendix~\ref{appendix:model_evaluation_prompts}.

%=====================================================
\begin{figure}
\centering
    \includegraphics[width=0.42\textwidth]{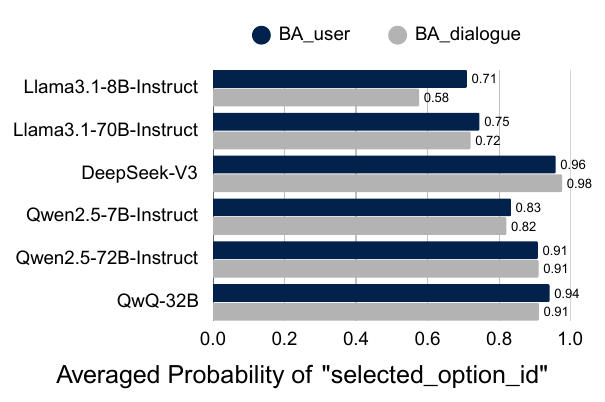}
    \caption{Mean probability of the “\texttt{selected\_option\_id}” in \texttt{BA\_user} and \texttt{BA\_dialogue}, reflecting model confidence. Most models show similar decisiveness across both scenarios.}
\label{fig:confidence_level}
\end{figure}

% +++++++++++++++++++++++++++++++++++++++++++++++++++++++++++++++
\begin{figure*}[t]
\centering
    \setlength{\tabcolsep}{4pt} % optional: shrink horizontal cell padding
        \begin{tabular}{@{}c@{\hspace{1em}}c@{}}
        \multicolumn{2}{c}{%
          \includegraphics[width=0.85\textwidth]{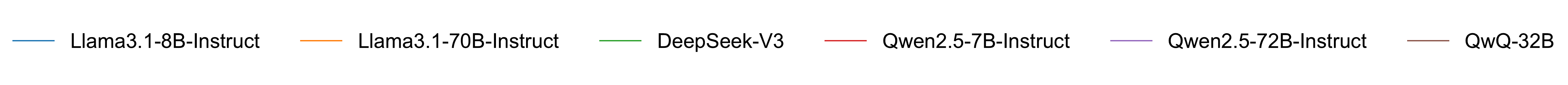}%
        } \\[-1ex]
        \subfloat[\texttt{BA\_user} grouped by ``age'']{%
          \includegraphics[width=0.37\textwidth]{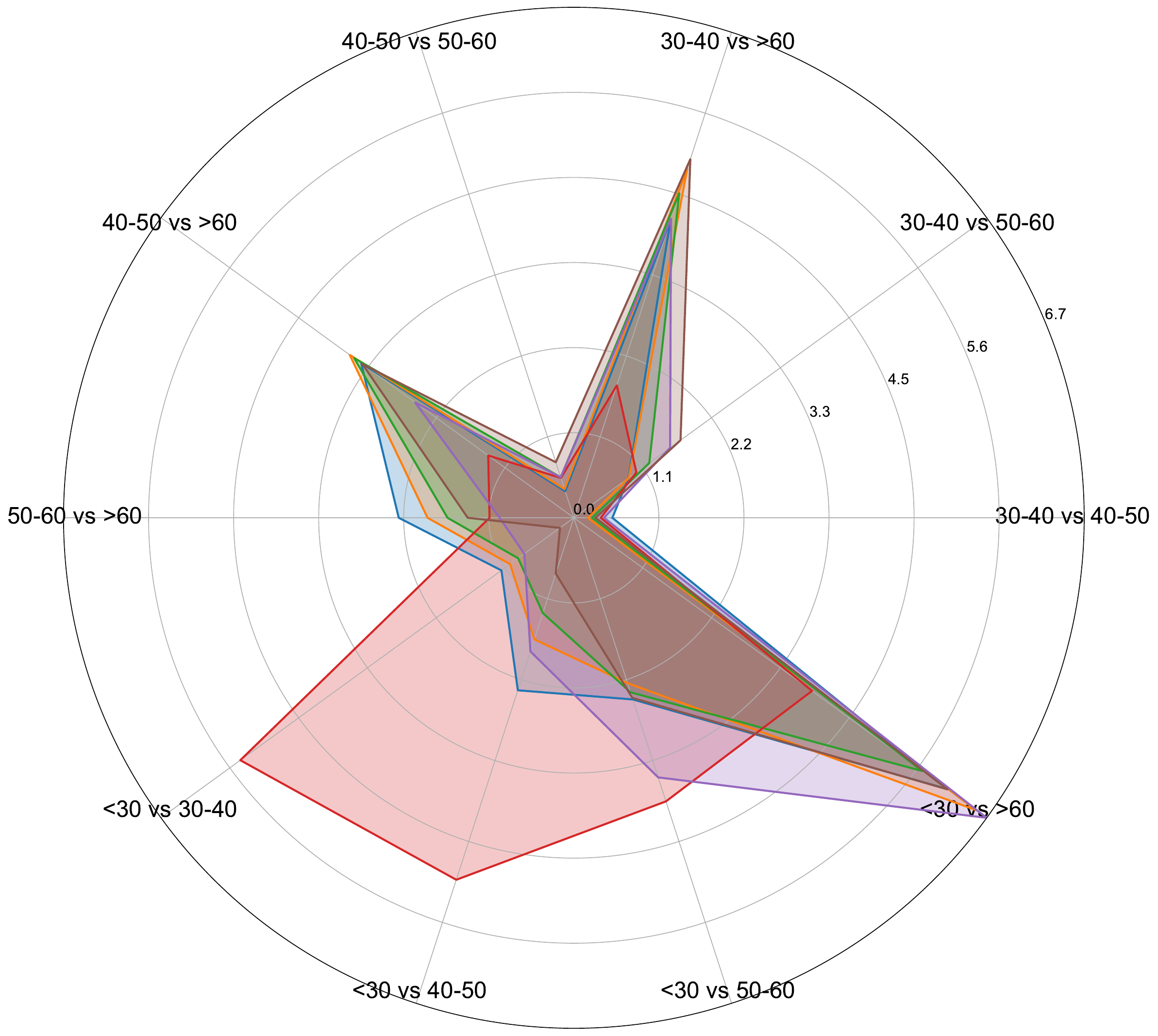}%
          \label{fig:BA_user_age}
        }
        &
        \subfloat[\texttt{BA\_dialogue} grouped by ``age'']{%
          \includegraphics[width=0.37\textwidth]{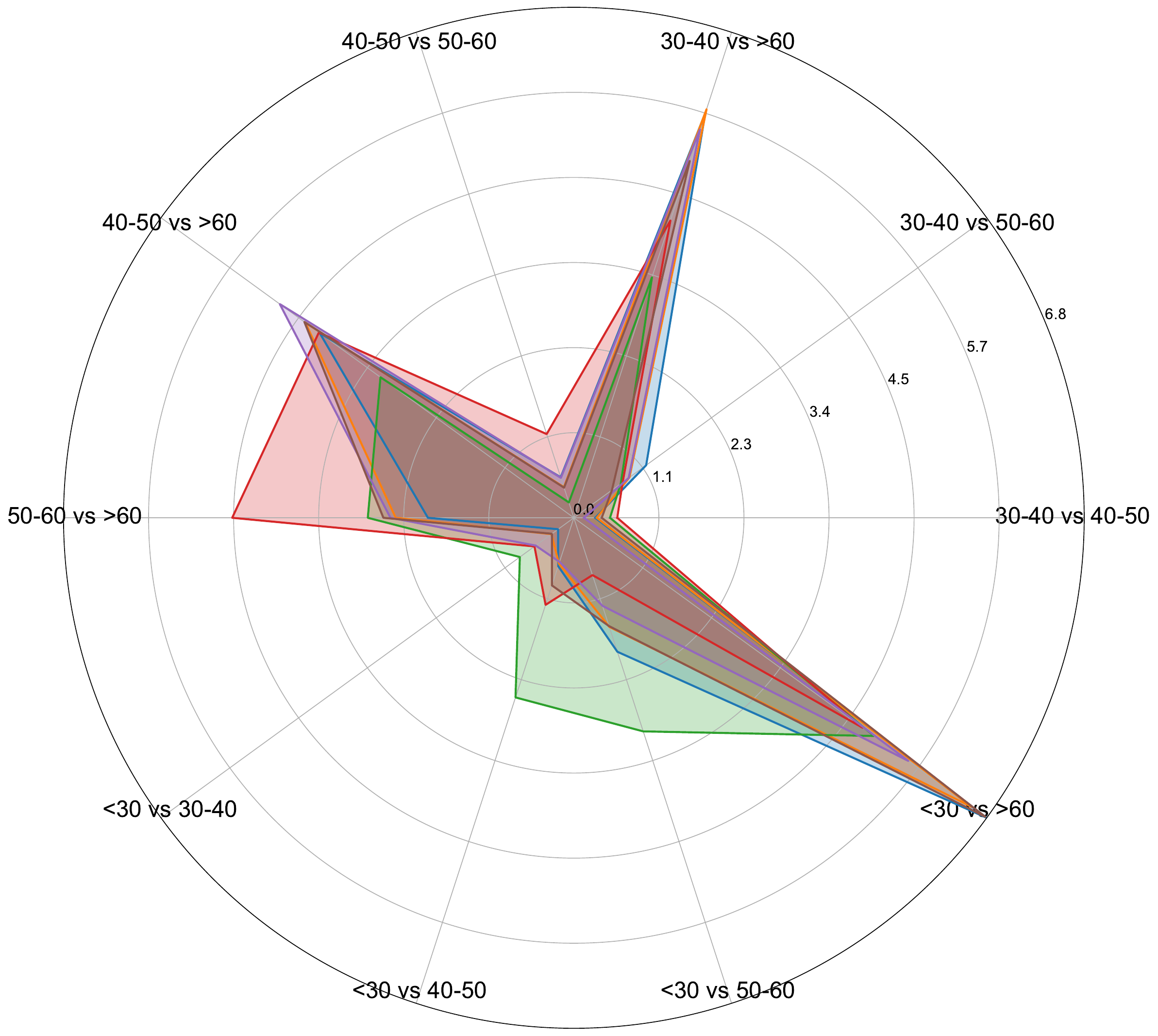}%
          \label{fig:BA_dialogue_age}
        } \\
        \subfloat[\texttt{BA\_user} grouped by ``education evel'']{\includegraphics[width=0.48\textwidth]{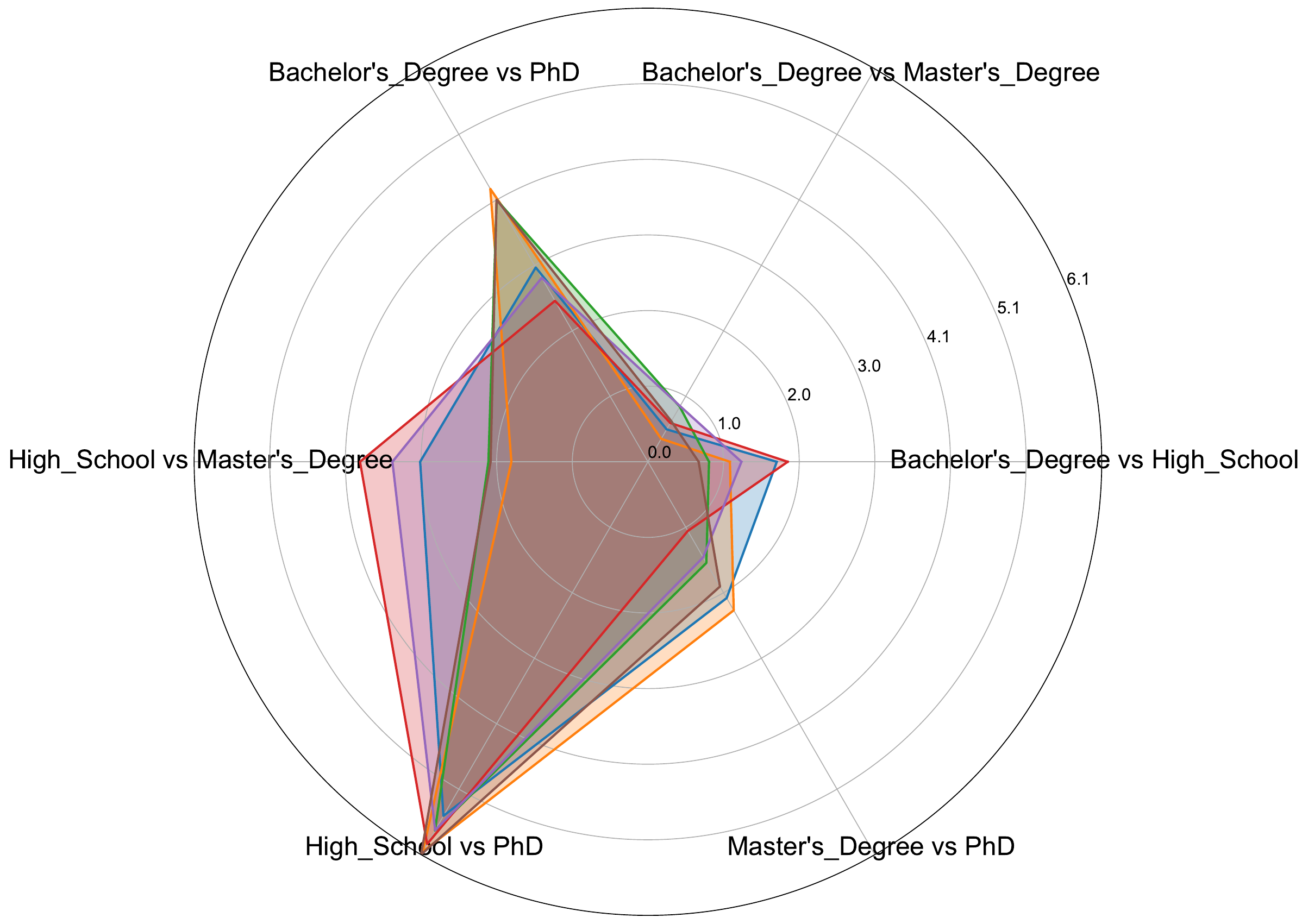}} 
        &
        \subfloat[\texttt{BA\_dialogue} grouped by ``education evel'']{\includegraphics[width=0.48\textwidth]{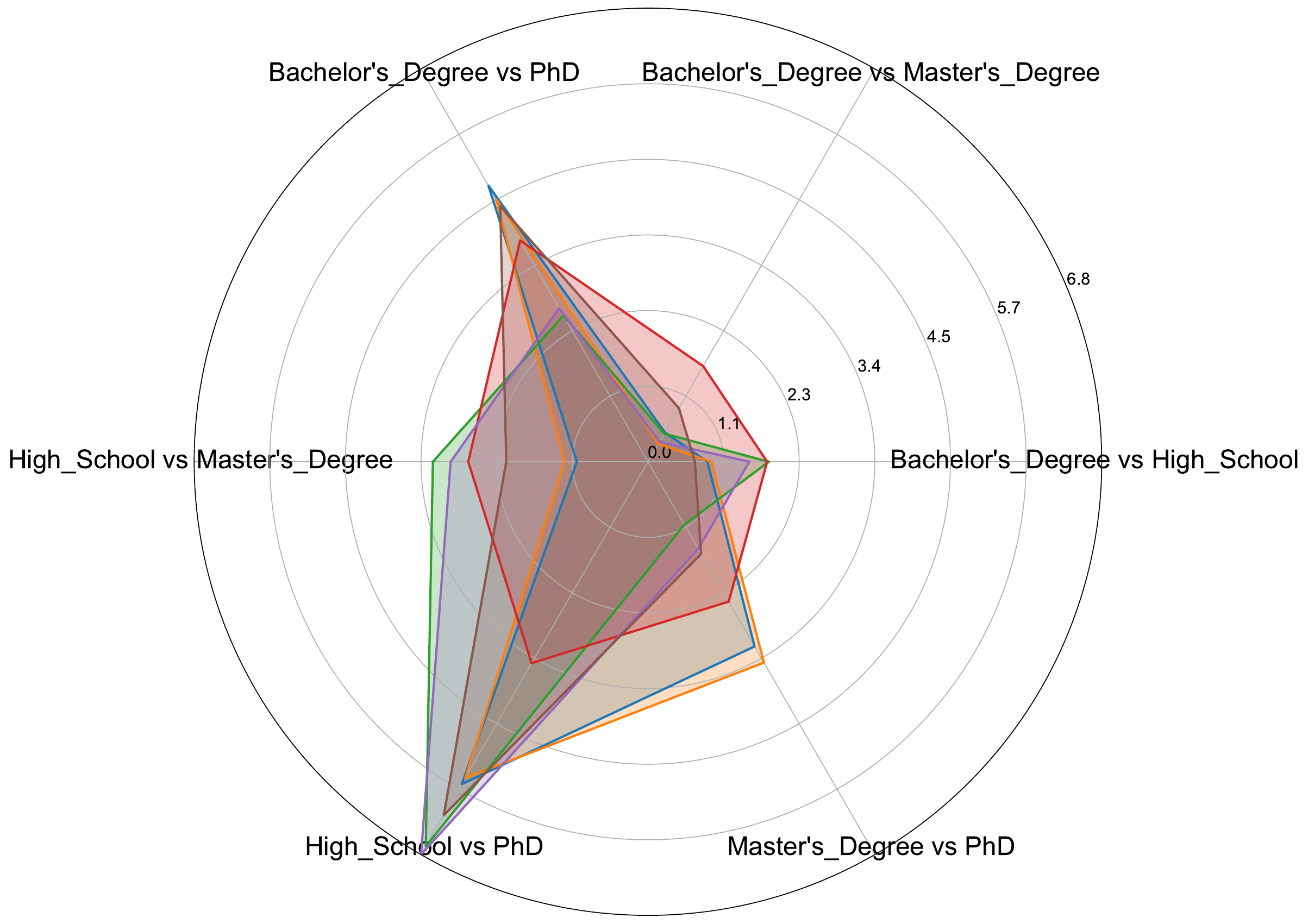}} 
    \end{tabular}
\caption{The measurement results for \texttt{BA\_user} and \texttt{BA\_dialogue} are shown below. The first row compares groups by ``Age,'' while the second row presents results for ``Education Level.'' Most models exhibit a positive correlation between computed distances and demographic differences. }
\label{fig:ba_evaluation_1}
\end{figure*}

%+++++++++++++++++++++++++++++++++++++++++++++++++++++++++++++++
\begin{table*}
\centering
\small
\begin{tabular}{l|ccc}
    \toprule
    \textbf{Models} & \textbf{Measured Distance $\downarrow$} & \textbf{Baseline} & \textbf{ \(\text{Distance} /\text{Baseline}\) $\downarrow$ } \\
    \midrule
    Llama3.1-8B-Instruct & 0.305 & 0.312 & 0.978 \\
    %\hline
    Llama3.1-70B-Instruct & 0.214 & 0.225 & 0.951 \\
    %\hline
    Qwen2.5-7B-Instruct & 0.276 & 0.276 & 1.000 \\
    %\hline
    Qwen2.5-72B-Instruct & 0.176 & 0.190 & 0.926 \\
    %\hline
    DeepSeek-V3 & 0.118 & 0.128 & 0.922 \\
    %\hline
    QwQ-32B & \textbf{0.112} & 0.125 & \textbf{0.896} \\
    \bottomrule
\end{tabular}
\caption{Results of evaluating LLMs' capability to maintain response consistency across different context formats. Both absolute distances and relative divergence ratios over the baseline are displayed in the table. \textit{QwQ-32B} has both the lowest absolute distance values and the lowest relative value of distance over baseline. }
\label{tab:consistency_results}
\end{table*}
%+++++++++++++++++++++++++++++++++++++++++++++++++++++++++++++++

\section{Evaluation Outcomes}

%\subsection{Model Confidence across Scenarios}

Recall that the selected ID is the one assigned the highest probability by the model. The probability value associated with this ID reflects the model’s confidence in its selection. We treat this probability as the confidence score for the selected ID and use the average of these scores across all responses to estimate the model’s overall confidence. 
%We assess each model's confidence by calculating the mean normalized probability of the \texttt{selected\_option\_id} across both \(R_U\) (\texttt{BA\_user}) and \(R_D\) (\texttt{BA\_dialogue}). 
As illustrated in Figure~\ref{fig:confidence_level}, all models—except \textit{Llama3.1-8B-Instruct}, which shows slightly reduced confidence in the dialogue setting—exhibit consistently high confidence across both contexts, supporting the interpretation that their selections reflect genuine preferences and reinforcing the validity of our subsequent analyses.

\subsection{Behavior Adaptation} 
%in \texttt{BA\_user} and \texttt{BA\_dialogue}}

Next, we assess behavior adaptation by varying demographic attributes. Responses in \(R_U\) and \(R_D\) are grouped by age, education, occupation, and nationality, and we compute inter-group divergences and corresponding baselines as specified in Table~\ref{tab:measure_forms}. Adaptation is quantified as the ratio of each divergence to its baseline (baseline=1). Detailed analyses for age and education appear below; further results are available in Appendix~\ref{appendix:more_evaluation_outcomes}.

\paratitle{Age:} We group responses \(R_U\) and \(R_D\) into 10-year brackets (\eg “30–40”), with “\(<30\)” and “\(>60\)” as boundary categories. Figure~\ref{fig:ba_evaluation_1} (first row) presents results for both \texttt{BA\_user} and \texttt{BA\_dialogue}. In both scenarios, models exhibit a positive correlation between age disparity and behavioral divergence (\eg maximum divergence between “\(<30\)” and “\(>60\)”). Notably, models exhibit greater consistency in the \texttt{BA\_dialogue} scenario than in \texttt{BA\_user}, indicating that age information conveyed through dialogue enables more stable adaptation across age groups.

\paratitle{Education Level:} Educational attainment provides a precise grouping criterion. Figure~\ref{fig:ba_evaluation_1} (second row) reports the divergences and baselines for both \texttt{BA\_user} and \texttt{BA\_dialogue}. As with age, models exhibit greater divergence in responses' values as educational disparity increases (e.g., highest divergence between high school and doctoral degrees), whether the attribute is provided explicitly or via dialogue. These results further demonstrate models’ capacity to incorporate educational background into value and nuance adaptation.

Additional evaluations using other attribute-based groupings are presented in Appendix~\ref{appendix:more_evaluation_outcomes}. These findings show that most LLMs accurately infer sociodemographic attributes and adjust their responses accordingly, with larger attribute differences producing greater shifts in expressed values. Crucially, despite dialogue history being less explicit than user profiles, models still adapt their behavior to align with user characteristics.

\subsection{Consistency across Context Formats}

% Given that models effectively adjust their responses to sociodemographic attributes provided via explicit profiles or multi-turn dialogues, 
We next evaluate each model’s behavioral consistency across context formats (scenario \texttt{Consistency}). Following Section~\ref{sec:distance_definition}, we compute, for each model, both the distance and its baseline. While absolute distances quantify response variability across formats, they may conflate behavioral alignment with format‐induced noise~\cite{he2024doespromptformattingimpact}. To disentangle these effects, we also report the ratio of measured distance to baseline: lower ratios signify stronger consistency in adapting to demographic cues across differing formats.

Table~\ref{tab:consistency_results} summarizes these metrics. Smaller tested models (with parameters smaller than 10B) exhibit both higher absolute EMDs and ratios near or equal to 1, indicating that prompt-format variability outweighs demographic alignment. In contrast, larger models achieve lower ratios, reflecting greater robustness to format changes and improved consistency in behavior adaptation. Furthermore, we observe a positive correlation between the measured consistency and benchmark performance on language understanding and reasoning tasks (e.g., MMLU Pro~\cite{mmlupro}, BigBench Hard~\cite{bigbenchhard}): models with superior cognitive capabilities tend to maintain higher consistency across formats.

The reasoning-augmented model \textit{QwQ-32B}, specifically trained through reinforcement learning for reasoning, achieves the highest consistency, despite its relatively smaller size. We attribute this superior consistency to its enhanced reasoning capabilities. To substantiate this, we recorded and analyzed its reasoning traces, revealing that \textit{QwQ-32B} systematically revisits and integrates the provided sociodemographic attributes, whether explicitly stated or inferred from dialogue history. This iterative process enables the model to select responses more precisely aligned with the given attributes, promoting consistency across context formats. The ability to methodically revisit and incorporate attribute details likely contributes to its superior alignment performance. Reasoning process samples are provided in Appendix~\ref{appendix:reasoning_samples}.

%In summary, performance in scenario \texttt{Consistency} depends on model capacity and reasoning ability. It highlights the 

\section{Conclusion}

This study presents a novel framework for evaluating how LLMs adapt their behavioral outputs when sociodemographic attributes are provided through two distinct interaction formats: (i) explicit single-turn prompts and (ii) multi-turn dialogue integration. We then systematically measure cross-format consistency to assess robust sociodemographic adaptation.
Our findings indicate that most models adjust effectively to single-format attribute changes, particularly in attributes like age and education level, with the degree of value adjustment positively correlated with the magnitude of attribute change. However, significant discrepancies arise in cross-format scenarios. Smaller models often struggle to maintain consistent alignment across formats, while larger, reasoning-augmented models demonstrate more stable performance. 
%Notably, the reasoning-enhanced \textit{QwQ-32B} model exhibited the highest consistency, underscoring the critical role of reasoning proficiency in achieving robust contextual adaptation across diverse interaction paradigms.

\section*{Limitations}
This study has a few limitations that require further investigation in future research. 
\begin{itemize}
    \item \textbf{Prompt Design:} Prompts for querying models to determine the most suitable response are designed based on our experience and experiments with sample cases, given the lack of established guidelines for optimizing prompts in value survey question answering. Future research could investigate how prompt content variations (besides the embedded information) impact models' behavior adaptation. 
   \item \textbf{Limited Scope of Value Survey:} The value survey used to assess model behavior consists of a relatively small number of questions from the VSM 2013 survey, which has been criticized for its simplicity and limited scope. However, it aligns well with our generated dataset, as it focuses on values reflected in career-related questions. Future studies could enhance the evaluation by incorporating larger and more diverse question sets.  
   \item \textbf{Single Source of Dialogue:} The evaluation of \texttt{BA\_dialogue} is based on dialogues generated by GPT-4O as the QA bot. When providing the tested models with the dialogue history, GPT’s specific response styles may influence the behavior of the tested models, which could be further explored in future work.

\end{itemize}

\section*{Ethical Considerations}
 
Our study examines how LLMs adapt outputs to users' sociodemographic contexts through explicit (profile‑based) and implicit (dialogue‑embedded) input formats. Misalignment or inconsistencies in model outputs can reinforce stereotypes or erode user trust in cross‑cultural interactions. To enable controlled, privacy‑preserving evaluations, we curate and open‑source a synthetic dataset, with all user profiles derived from a publicly available synthetic dataset (see Section~\ref{sec:dataset_generation}), thereby eliminating any risk of personal identifying information (PII) exposure or consent violations. However, given the dataset's controlled topical focus, excessive reliance on these examples without careful consideration of their suitability for use cases may introduce bias and overlook the broader diversity of real‑world interactions.

\bibliography{refs}

\begin{thebibliography}{50}
\providecommand{\natexlab}[1]{#1}

\bibitem[{Abdullin et~al.(2024)Abdullin, Molla-Aliod, Ofoghi, Yearwood, and Li}]{abdullin2024syntheticdialoguedatasetgeneration}
Yelaman Abdullin, Diego Molla-Aliod, Bahadorreza Ofoghi, John Yearwood, and Qingyang Li. 2024.
\newblock \href {https://arxiv.org/abs/2401.17461} {Synthetic dialogue dataset generation using llm agents}.
\newblock \emph{Preprint}, arXiv:2401.17461.

\bibitem[{Altenburger et~al.(2024)Altenburger, Jiang, Kraut, Wang, and Dwivedi-Yu}]{altenburger2024examiningrolerelationshipalignment}
Kristen~M. Altenburger, Hongda Jiang, Robert~E. Kraut, Yi-Chia Wang, and Jane Dwivedi-Yu. 2024.
\newblock \href {https://arxiv.org/abs/2410.01708} {Examining the role of relationship alignment in large language models}.
\newblock \emph{Preprint}, arXiv:2410.01708.

\bibitem[{Arora et~al.(2023)Arora, Kaffee, and Augenstein}]{arora2023probing}
Arnav Arora, Lucie-Aimée Kaffee, and Isabelle Augenstein. 2023.
\newblock \href {https://arxiv.org/abs/2203.13722} {Probing pre-trained language models for cross-cultural differences in values}.
\newblock \emph{Preprint}, arXiv:2203.13722.

\bibitem[{Chen et~al.(2024)Chen, Chen, Zhou, Yishen, and He}]{chen-etal-2024-diahalu}
Kedi Chen, Qin Chen, Jie Zhou, He~Yishen, and Liang He. 2024.
\newblock \href {https://doi.org/10.18653/v1/2024.findings-emnlp.529} {{D}ia{H}alu: A dialogue-level hallucination evaluation benchmark for large language models}.
\newblock In \emph{Findings of the Association for Computational Linguistics: EMNLP 2024}, pages 9057--9079, Miami, Florida, USA. Association for Computational Linguistics.

\bibitem[{Dam et~al.(2024)Dam, Hong, Qiao, and Zhang}]{dam2024completesurveyllmbasedai}
Sumit~Kumar Dam, Choong~Seon Hong, Yu~Qiao, and Chaoning Zhang. 2024.
\newblock \href {https://arxiv.org/abs/2406.16937} {A complete survey on llm-based ai chatbots}.
\newblock \emph{Preprint}, arXiv:2406.16937.

\bibitem[{DeepSeek-AI(2024)}]{deepseekai2024deepseekv3technicalreport}
DeepSeek-AI. 2024.
\newblock \href {https://arxiv.org/abs/2412.19437} {Deepseek-v3 technical report}.
\newblock \emph{Preprint}, arXiv:2412.19437.

\bibitem[{Dong et~al.(2024)Dong, Ruan, Cai, Lai, Xu, Zhao, and Chen}]{dong2024xgrammarflexibleefficientstructured}
Yixin Dong, Charlie~F. Ruan, Yaxing Cai, Ruihang Lai, Ziyi Xu, Yilong Zhao, and Tianqi Chen. 2024.
\newblock \href {https://arxiv.org/abs/2411.15100} {Xgrammar: Flexible and efficient structured generation engine for large language models}.
\newblock \emph{Preprint}, arXiv:2411.15100.

\bibitem[{Dubey and et~al.(2024)}]{dubey2024llama3herdmodels}
Abhimanyu Dubey and Abhinav~Jauhri et~al. 2024.
\newblock \href {https://arxiv.org/abs/2407.21783} {The llama 3 herd of models}.
\newblock \emph{Preprint}, arXiv:2407.21783.

\bibitem[{DURMUS et~al.(2024)DURMUS, Nguyen, Liao, Schiefer, Askell, Bakhtin, Chen, Hatfield-Dodds, Hernandez, Joseph, Lovitt, McCandlish, Sikder, Tamkin, Thamkul, Kaplan, Clark, and Ganguli}]{durmus2024towards}
Esin DURMUS, Karina Nguyen, Thomas Liao, Nicholas Schiefer, Amanda Askell, Anton Bakhtin, Carol Chen, Zac Hatfield-Dodds, Danny Hernandez, Nicholas Joseph, Liane Lovitt, Sam McCandlish, Orowa Sikder, Alex Tamkin, Janel Thamkul, Jared Kaplan, Jack Clark, and Deep Ganguli. 2024.
\newblock \href {https://openreview.net/forum?id=zl16jLb91v} {Towards measuring the representation of subjective global opinions in language models}.
\newblock In \emph{First Conference on Language Modeling}.

\bibitem[{Freedman et~al.(2007)Freedman, Pisani, and Purves}]{freedman2007statistics}
David Freedman, Robert Pisani, and Roger Purves. 2007.
\newblock Statistics (international student edition).
\newblock \emph{Pisani, R. Purves, 4th edn. WW Norton \& Company, New York}.

\bibitem[{Fung et~al.(2016)Fung, Ho, Zhang, Zhang, Noels, and Tam}]{Fung2016-re}
Helene~H Fung, Yuan~Wan Ho, Rui Zhang, Xin Zhang, Kimberly~A Noels, and Kim-Pong Tam. 2016.
\newblock Age differences in personal values: Universal or cultural specific?
\newblock \emph{Psychol. Aging}, 31(3):274--286.

\bibitem[{Gao et~al.(2023)Gao, Lian, Zhou, Fu, and Wang}]{gao-etal-2023-livechat}
Jingsheng Gao, Yixin Lian, Ziyi Zhou, Yuzhuo Fu, and Baoyuan Wang. 2023.
\newblock \href {https://doi.org/10.18653/v1/2023.acl-long.858} {{L}ive{C}hat: A large-scale personalized dialogue dataset automatically constructed from live streaming}.
\newblock In \emph{Proceedings of the 61st Annual Meeting of the Association for Computational Linguistics (Volume 1: Long Papers)}, pages 15387--15405, Toronto, Canada. Association for Computational Linguistics.

\bibitem[{Gelfand and Raver(2011)}]{Gelfand2011-ei}
Michele~J Gelfand and Jana L et~al. Raver. 2011.
\newblock Differences between tight and loose cultures: a 33-nation study.
\newblock \emph{Science}, 332(6033):1100--1104.

\bibitem[{Gerlach and Eriksson(2021)}]{Gerlach2021-ww}
Philipp Gerlach and Kimmo Eriksson. 2021.
\newblock Measuring cultural dimensions: External validity and internal consistency of hofstede's {VSM} 2013 scales.
\newblock \emph{Front. Psychol.}, 12:662604.

\bibitem[{Gu et~al.(2025)Gu, Jiang, Shi, Tan, Zhai, Xu, Li, Shen, Ma, Liu, Wang, Zhang, Wang, Gao, Ni, and Guo}]{gu2025surveyllmasajudge}
Jiawei Gu, Xuhui Jiang, Zhichao Shi, Hexiang Tan, Xuehao Zhai, Chengjin Xu, Wei Li, Yinghan Shen, Shengjie Ma, Honghao Liu, Saizhuo Wang, Kun Zhang, Yuanzhuo Wang, Wen Gao, Lionel Ni, and Jian Guo. 2025.
\newblock \href {https://arxiv.org/abs/2411.15594} {A survey on llm-as-a-judge}.
\newblock \emph{Preprint}, arXiv:2411.15594.

\bibitem[{Gupta et~al.(2024)Gupta, Sheth, Raina, Gales, and Fritz}]{gupta-etal-2024-llm}
Akash Gupta, Ivaxi Sheth, Vyas Raina, Mark Gales, and Mario Fritz. 2024.
\newblock \href {https://doi.org/10.18653/v1/2024.emnlp-main.811} {{LLM} task interference: An initial study on the impact of task-switch in conversational history}.
\newblock In \emph{Proceedings of the 2024 Conference on Empirical Methods in Natural Language Processing}, pages 14633--14652, Miami, Florida, USA. Association for Computational Linguistics.

\bibitem[{Haerpfer et~al.(2020)Haerpfer, Inglehart, Moreno, Welzel, Kizilova, Diez-Medrano, Lagos, and Norris}]{Haerpfer2020-qe}
C~Haerpfer, R~Inglehart, A~Moreno, C~Welzel, K~Kizilova, J~Diez-Medrano, M~Lagos, and P~Norris. 2020.
\newblock \emph{World Values Survey: Round Seven - {Country-Pooled} Datafile}.
\newblock JD Systems Institute \& WVSA Secretariat, Madrid, Spain \& Vienna, Austria.

\bibitem[{He et~al.(2024)He, Rungta, Koleczek, Sekhon, Wang, and Hasan}]{he2024doespromptformattingimpact}
Jia He, Mukund Rungta, David Koleczek, Arshdeep Sekhon, Franklin~X Wang, and Sadid Hasan. 2024.
\newblock \href {https://arxiv.org/abs/2411.10541} {Does prompt formatting have any impact on llm performance?}
\newblock \emph{Preprint}, arXiv:2411.10541.

\bibitem[{Hoehler(2000)}]{Hoehler2000BiasAP}
Fred~K. Hoehler. 2000.
\newblock \href {https://api.semanticscholar.org/CorpusID:43684174} {Bias and prevalence effects on kappa viewed in terms of sensitivity and specificity.}
\newblock \emph{Journal of clinical epidemiology}, 53 5:499--503.

\bibitem[{Hofstede and Hofstede(2016)}]{vsm2013-jj}
G~Hofstede and G.~J. Hofstede. 2016.
\newblock {VSM} 2013.
\newblock \url{https://geerthofstede.com/research-and-vsm/vsm-2013/}.
\newblock Accessed: 2024-1-11.

\bibitem[{Jandaghi et~al.(2023)Jandaghi, Sheng, Bai, Pujara, and Sidahmed}]{faithfulPersonaChat}
Pegah Jandaghi, XiangHai Sheng, Xinyi Bai, Jay Pujara, and Hakim Sidahmed. 2023.
\newblock \href {https://arxiv.org/abs/2312.10007} {Faithful persona-based conversational dataset generation with large language models}.
\newblock \emph{Preprint}, arXiv:2312.10007.

\bibitem[{Kharchenko et~al.(2024)Kharchenko, Roosta, Chadha, and Shah}]{kharchenko2024llmsrepresentvaluescultures}
Julia Kharchenko, Tanya Roosta, Aman Chadha, and Chirag Shah. 2024.
\newblock \href {https://arxiv.org/abs/2406.14805} {How well do llms represent values across cultures? empirical analysis of llm responses based on hofstede cultural dimensions}.
\newblock \emph{Preprint}, arXiv:2406.14805.

\bibitem[{Kovač et~al.(2023)Kovač, Sawayama, Portelas, Colas, Dominey, and Oudeyer}]{kovač2023large}
Grgur Kovač, Masataka Sawayama, Rémy Portelas, Cédric Colas, Peter~Ford Dominey, and Pierre-Yves Oudeyer. 2023.
\newblock \href {https://arxiv.org/abs/2307.07870} {Large language models as superpositions of cultural perspectives}.
\newblock \emph{Preprint}, arXiv:2307.07870.

\bibitem[{Kwon et~al.(2023)Kwon, Li, Zhuang, Sheng, Zheng, Yu, Gonzalez, Zhang, and Stoica}]{vllm}
Woosuk Kwon, Zhuohan Li, Siyuan Zhuang, Ying Sheng, Lianmin Zheng, Cody~Hao Yu, Joseph~E. Gonzalez, Hao Zhang, and Ion Stoica. 2023.
\newblock Efficient memory management for large language model serving with pagedattention.
\newblock In \emph{Proceedings of the ACM SIGOPS 29th Symposium on Operating Systems Principles}.

\bibitem[{Leng(2023)}]{leng_llm_2023}
Quinn Leng. 2023.
\newblock Best practices for {LLM} evaluation of {RAG} applications: A case study on the databricks documentation bot.
\newblock \url{https://www.databricks.com/blog/LLM-auto-eval-best-practices-RAG}.
\newblock Accessed: 2025-04-13.

\bibitem[{Lewis et~al.(2019)Lewis, Liu, Goyal, Ghazvininejad, Mohamed, Levy, Stoyanov, and Zettlemoyer}]{bart-large}
Mike Lewis, Yinhan Liu, Naman Goyal, Marjan Ghazvininejad, Abdelrahman Mohamed, Omer Levy, Veselin Stoyanov, and Luke Zettlemoyer. 2019.
\newblock \href {https://arxiv.org/abs/1910.13461} {{BART:} denoising sequence-to-sequence pre-training for natural language generation, translation, and comprehension}.
\newblock \emph{CoRR}, abs/1910.13461.

\bibitem[{Li et~al.(2024)Li, Chen, Zhang, Lou, Li, Sun, Liu, and Liu}]{li2024benchmarkingbiaslargelanguage}
Xinyue Li, Zhenpeng Chen, Jie~M. Zhang, Yiling Lou, Tianlin Li, Weisong Sun, Yang Liu, and Xuanzhe Liu. 2024.
\newblock \href {https://arxiv.org/abs/2411.00585} {Benchmarking bias in large language models during role-playing}.
\newblock \emph{Preprint}, arXiv:2411.00585.

\bibitem[{Liu et~al.(2025)Liu, Dey, Zhao, tse Huang, Gupta, Liu, and Zhao}]{liu2025llmsgraspimplicitcultural}
Ziyi Liu, Priyanka Dey, Zhenyu Zhao, Jen tse Huang, Rahul Gupta, Yang Liu, and Jieyu Zhao. 2025.
\newblock \href {https://arxiv.org/abs/2504.01127} {Can llms grasp implicit cultural values? benchmarking llms' metacognitive cultural intelligence with cq-bench}.
\newblock \emph{Preprint}, arXiv:2504.01127.

\bibitem[{Lomazzi and Seddig(2020)}]{genderandvalues}
Vera Lomazzi and Daniel Seddig. 2020.
\newblock \href {https://doi.org/10.1177/1069397120915454} {Gender role attitudes in the international social survey programme: Cross-national comparability and relationships to cultural values}.
\newblock \emph{Cross-Cultural Research}, 54:106939712091545.

\bibitem[{Masoud et~al.(2024)Masoud, Liu, Ferianc, Treleaven, and Rodrigues}]{masoud2024culturalalignmentlargelanguage}
Reem~I. Masoud, Ziquan Liu, Martin Ferianc, Philip Treleaven, and Miguel Rodrigues. 2024.
\newblock \href {https://arxiv.org/abs/2309.12342} {Cultural alignment in large language models: An explanatory analysis based on hofstede's cultural dimensions}.
\newblock \emph{Preprint}, arXiv:2309.12342.

\bibitem[{Moore et~al.(2024)Moore, Deshpande, and Yang}]{moore-etal-2024-large}
Jared Moore, Tanvi Deshpande, and Diyi Yang. 2024.
\newblock \href {https://doi.org/10.18653/v1/2024.findings-emnlp.891} {Are large language models consistent over value-laden questions?}
\newblock In \emph{Findings of the Association for Computational Linguistics: EMNLP 2024}, pages 15185--15221, Miami, Florida, USA. Association for Computational Linguistics.

\bibitem[{Nielsen(2020)}]{Jensen_Shannon_Centroid}
Frank Nielsen. 2020.
\newblock \href {https://doi.org/10.3390/e22020221} {On a generalization of the jensen–shannon divergence and the jensen–shannon centroid}.
\newblock \emph{Entropy}, 22(2).

\bibitem[{OpenAI et~al.(2024)OpenAI, Achiam, and et~al.}]{openai2024gpt4technicalreport}
OpenAI, Josh Achiam, and Steven~Adler et~al. 2024.
\newblock \href {https://arxiv.org/abs/2303.08774} {Gpt-4 technical report}.
\newblock \emph{Preprint}, arXiv:2303.08774.

\bibitem[{Qian et~al.(2021)Qian, Li, Zhong, Guo, Ma, Zhu, Liu, Dou, and Wen}]{pchatbot}
Hongjin Qian, Xiaohe Li, Hanxun Zhong, Yu~Guo, Yueyuan Ma, Yutao Zhu, Zhanliang Liu, Zhicheng Dou, and Ji-Rong Wen. 2021.
\newblock \href {https://doi.org/10.1145/3404835.3463239} {Pchatbot: A large-scale dataset for personalized chatbot}.
\newblock In \emph{Proceedings of the 44th International ACM SIGIR Conference on Research and Development in Information Retrieval}, SIGIR '21, page 2470–2477, New York, NY, USA. Association for Computing Machinery.

\bibitem[{Qwen et~al.(2025)Qwen, :, and et~al.}]{qwen2025qwen25technicalreport}
Qwen, :, and An~Yang et~al. 2025.
\newblock \href {https://arxiv.org/abs/2412.15115} {Qwen2.5 technical report}.
\newblock \emph{Preprint}, arXiv:2412.15115.

\bibitem[{Rubner et~al.(1998)Rubner, Tomasi, and Guibas}]{EMD}
Y.~Rubner, C.~Tomasi, and L.J. Guibas. 1998.
\newblock \href {https://doi.org/10.1109/ICCV.1998.710701} {A metric for distributions with applications to image databases}.
\newblock In \emph{Sixth International Conference on Computer Vision (IEEE Cat. No.98CH36271)}, pages 59--66.

\bibitem[{Shrout and Fleiss(1979)}]{Shrout1979-ICC}
P~E Shrout and J~L Fleiss. 1979.
\newblock Intraclass correlations: uses in assessing rater reliability.
\newblock \emph{Psychol. Bull.}, 86(2):420--428.

\bibitem[{Sicilia et~al.(2024)Sicilia, Gates, and Alikhani}]{sicilia2024humbelhumanintheloopapproachevaluating}
Anthony Sicilia, Jennifer~C. Gates, and Malihe Alikhani. 2024.
\newblock \href {https://arxiv.org/abs/2305.14195} {Humbel: A human-in-the-loop approach for evaluating demographic factors of language models in human-machine conversations}.
\newblock \emph{Preprint}, arXiv:2305.14195.

\bibitem[{Sukiennik et~al.(2025)Sukiennik, Gao, Xu, and Li}]{sukiennik2025evaluationculturalvaluealignment}
Nicholas Sukiennik, Chen Gao, Fengli Xu, and Yong Li. 2025.
\newblock \href {https://arxiv.org/abs/2504.08863} {An evaluation of cultural value alignment in llm}.
\newblock \emph{Preprint}, arXiv:2504.08863.

\bibitem[{Suzgun et~al.(2022)Suzgun, Scales, Schärli, Gehrmann, Tay, Chung, Chowdhery, Le, Chi, Zhou, and Wei}]{bigbenchhard}
Mirac Suzgun, Nathan Scales, Nathanael Schärli, Sebastian Gehrmann, Yi~Tay, Hyung~Won Chung, Aakanksha Chowdhery, Quoc~V. Le, Ed~H. Chi, Denny Zhou, and Jason Wei. 2022.
\newblock \href {https://arxiv.org/abs/2210.09261} {Challenging big-bench tasks and whether chain-of-thought can solve them}.
\newblock \emph{Preprint}, arXiv:2210.09261.

\bibitem[{Twenge(2017)}]{twenge2017smartphones}
Jean~M. Twenge. 2017.
\newblock Have smartphones destroyed a generation?
\newblock \emph{The Atlantic}.

\bibitem[{Wang et~al.(2024)Wang, Ma, Zhang, Ni, Chandra, Guo, Ren, Arulraj, He, Jiang, Li, Ku, Wang, Zhuang, Fan, Yue, and Chen}]{mmlupro}
Yubo Wang, Xueguang Ma, Ge~Zhang, Yuansheng Ni, Abhranil Chandra, Shiguang Guo, Weiming Ren, Aaran Arulraj, Xuan He, Ziyan Jiang, Tianle Li, Max Ku, Kai Wang, Alex Zhuang, Rongqi Fan, Xiang Yue, and Wenhu Chen. 2024.
\newblock \href {https://arxiv.org/abs/2406.01574} {Mmlu-pro: A more robust and challenging multi-task language understanding benchmark}.
\newblock \emph{Preprint}, arXiv:2406.01574.

\bibitem[{Williams et~al.(2018)Williams, Nangia, and Bowman}]{MultiNLI}
Adina Williams, Nikita Nangia, and Samuel Bowman. 2018.
\newblock \href {http://aclweb.org/anthology/N18-1101} {A broad-coverage challenge corpus for sentence understanding through inference}.
\newblock In \emph{Proceedings of the 2018 Conference of the North American Chapter of the Association for Computational Linguistics: Human Language Technologies, Volume 1 (Long Papers)}, pages 1112--1122. Association for Computational Linguistics.

\bibitem[{Wolf et~al.(2020)Wolf, Debut, Sanh, Chaumond, Delangue, Moi, Cistac, Rault, Louf, Funtowicz, Davison, Shleifer, von Platen, Ma, Jernite, Plu, Xu, Scao, Gugger, Drame, Lhoest, and Rush}]{wolf2020huggingfaces}
Thomas Wolf, Lysandre Debut, Victor Sanh, Julien Chaumond, Clement Delangue, Anthony Moi, Pierric Cistac, Tim Rault, Rémi Louf, Morgan Funtowicz, Joe Davison, Sam Shleifer, Patrick von Platen, Clara Ma, Yacine Jernite, Julien Plu, Canwen Xu, Teven~Le Scao, Sylvain Gugger, Mariama Drame, Quentin Lhoest, and Alexander~M. Rush. 2020.
\newblock \href {https://arxiv.org/abs/1910.03771} {Huggingface's transformers: State-of-the-art natural language processing}.
\newblock \emph{Preprint}, arXiv:1910.03771.

\bibitem[{Yamashita et~al.(2023)Yamashita, Inoue, Guo, Mochizuki, Kawahara, and Higashinaka}]{realpersonachat}
Sanae Yamashita, Koji Inoue, Ao~Guo, Shota Mochizuki, Tatsuya Kawahara, and Ryuichiro Higashinaka. 2023.
\newblock \href {https://aclanthology.org/2023.paclic-1.85/} {{R}eal{P}ersona{C}hat: A realistic persona chat corpus with interlocutors' own personalities}.
\newblock In \emph{Proceedings of the 37th Pacific Asia Conference on Language, Information and Computation}, pages 852--861, Hong Kong, China. Association for Computational Linguistics.

\bibitem[{Yao et~al.(2024)Yao, Yi, and Xie}]{yao2024claveadaptiveframeworkevaluating}
Jing Yao, Xiaoyuan Yi, and Xing Xie. 2024.
\newblock \href {https://arxiv.org/abs/2407.10725} {Clave: An adaptive framework for evaluating values of llm generated responses}.
\newblock \emph{Preprint}, arXiv:2407.10725.

\bibitem[{Zhang et~al.(2018)Zhang, Dinan, Urbanek, Szlam, Kiela, and Weston}]{personachat}
Saizheng Zhang, Emily Dinan, Jack Urbanek, Arthur Szlam, Douwe Kiela, and Jason Weston. 2018.
\newblock \href {https://doi.org/10.18653/v1/P18-1205} {Personalizing dialogue agents: {I} have a dog, do you have pets too?}
\newblock In \emph{Proceedings of the 56th Annual Meeting of the Association for Computational Linguistics (Volume 1: Long Papers)}, pages 2204--2213, Melbourne, Australia. Association for Computational Linguistics.

\bibitem[{Zhang et~al.(2024)Zhang, Zhang, Liu, Qi, Rong, Zhu, Cui, and Yang}]{zhang2024heterogeneousvaluealignmentevaluation}
Zhaowei Zhang, Ceyao Zhang, Nian Liu, Siyuan Qi, Ziqi Rong, Song-Chun Zhu, Shuguang Cui, and Yaodong Yang. 2024.
\newblock \href {https://arxiv.org/abs/2305.17147} {Heterogeneous value alignment evaluation for large language models}.
\newblock \emph{Preprint}, arXiv:2305.17147.

\bibitem[{Zhao et~al.(2024)Zhao, Mondal, Tandon, Dillion, Gray, and Gu}]{zhao2024worldvaluesbenchlargescalebenchmarkdataset}
Wenlong Zhao, Debanjan Mondal, Niket Tandon, Danica Dillion, Kurt Gray, and Yuling Gu. 2024.
\newblock \href {https://arxiv.org/abs/2404.16308} {Worldvaluesbench: A large-scale benchmark dataset for multi-cultural value awareness of language models}.
\newblock \emph{Preprint}, arXiv:2404.16308.

\bibitem[{Zheng et~al.(2023)Zheng, Chiang, Sheng, Zhuang, Wu, Zhuang, Lin, Li, Li, Xing, Zhang, Gonzalez, and Stoica}]{judgingllmasajudge}
Lianmin Zheng, Wei-Lin Chiang, Ying Sheng, Siyuan Zhuang, Zhanghao Wu, Yonghao Zhuang, Zi~Lin, Zhuohan Li, Dacheng Li, Eric~P. Xing, Hao Zhang, Joseph~E. Gonzalez, and Ion Stoica. 2023.
\newblock Judging llm-as-a-judge with mt-bench and chatbot arena.
\newblock In \emph{Proceedings of the 37th International Conference on Neural Information Processing Systems}, NIPS '23, Red Hook, NY, USA. Curran Associates Inc.

\end{thebibliography}

\appendix

\clearpage

%==========================================================
\begin{table*}
\centering
\small
\begin{tabular}{l|c|c|c|c|c}
    \toprule
    \multirow{3}{*}{\textbf{Prompt Type}} & \multirow{3}{*}{\textbf{Models}} & \multicolumn{4}{c}{\textbf{Hit Rate}} \\
    \cmidrule{3-6}
    & & \multicolumn{2}{c|}{vs \textbf{Ground Truth}} & \multicolumn{2}{c}{vs \textbf{gpt-4o-mini-2024-07-18}} \\
    \cmidrule{3-6}
    & & \text{~~~~~User1~~~~~} & \text{~~~~~User2~~~~~} & \text{~~~~~User1~~~~~} & \text{~~~~~User2~~~~} \\
    \midrule
    \multirow{2}{*}{\textbf{Zero-shot}} & Llama3-8B-Instruct  & 0.826 & 0.864 & 0.732 & 0.784  \\
    & Llama3.1-8B-Instruct & 0.821 & 0.864 & 0.731 & 0.796 \\
   
    \midrule
    \multirow{2}{*}{\textbf{One-shot}} & Llama3-8B-Instruct  & 0.849 & 0.895 & 0.775 & 0.839 \\
    & Llama3.1-8B-Instruct & 0.837 & 0.888 & 0.756 & 0.827 \\
    
    \midrule
    \multirow{2}{*}{\textbf{Five-shot}} & Llama3-8B-Instruct  & 0.851 & 0.906 & 0.784 & 0.854 \\
    & Llama3.1-8B-Instruct & 0.848 & 0.905 & 0.774 & 0.853  \\
    
    \midrule
    \multirow{2}{*}{\textbf{Ten-shot}} & Llama3-8B-Instruct  & 0.856 & 0.910 & 0.783 & 0.855 \\
    & Llama3.1-8B-Instruct & 0.852 & 0.905 & 0.775 & 0.853 \\
    
    \midrule
    Random & -- &0.5 & 0.5 & 0.18 & 0.19 \\
    \bottomrule
\end{tabular}
\caption{Recognition accuracy of persona attributes from dialogue by LLMs. We evaluate both ``Llama3.1-8B-Instruct'' and ``Llama3-8B-Instruct'', which belong to the same Llama model family. Model scores are compared against the expected accuracy of randomly selecting one correct attribute from the list of 10 candidates.}
\label{tab:persona_recognition_result}
\end{table*}
%==========================================================
\section{Capability of LLMs for Recognizing Persona Attributes}  
\label{appendix:persona_recognition}
As a preliminary stage of this study, we select \texttt{Llama3.1-8B-Instruct} as a representative model to verify its ability to extract user attributes from dialogue. This model is one of the smallest evaluated in our study. For evaluation, we utilize the FaithfulPersonaChat~\cite{faithfulPersonaChat} synthetic dataset, which comprises 5,648 unique synthetic personas and 11,001 conversations. Each conversation takes place between two users (User1 and User2), and the persona profiles of both participants are provided alongside the dialogue. Every synthetic user is assigned five persona attributes, with at least one of these attributes explicitly mentioned in their dialogues.

Given the dataset structure, we prompt the tested model with a conversation and ask it to identify one mentioned persona attribute for each user (evaluated one at a time). The model selects from a list of ten candidate attributes, which includes the user's five correct attributes and five randomly sampled attributes from other users. The selected attribute is then evaluated against two reference settings: (1) the ground truth, i.e., the user's complete set of five correct attributes, and (2) the output from \texttt{gpt-4o-mini}~\cite{openai2024gpt4technicalreport}, which is prompted to extract all identifiable persona attributes from the conversation. If the selected attribute appears in either reference, the model receives a score of 1; otherwise, the score is 0. The final score is calculated by aggregating the hit rate across all 11,001 conversations for each user. We test the model under different prompting strategies, ranging from zero-shot to ten-shot settings.

The results, shown in Table~\ref{tab:persona_recognition_result}, confirm that the model can \textit{effectively identify at least one persona attribute} from dialogues. This validation supports our main research objective: to investigate whether models can adapt their responses based on the recognized attributes.

\section{Prompts Design for  Dataset Generation Agent}
\label{appendix:dataset_generation_details}

\begin{figure*}
\centering
    \includegraphics[width=0.8\textwidth]{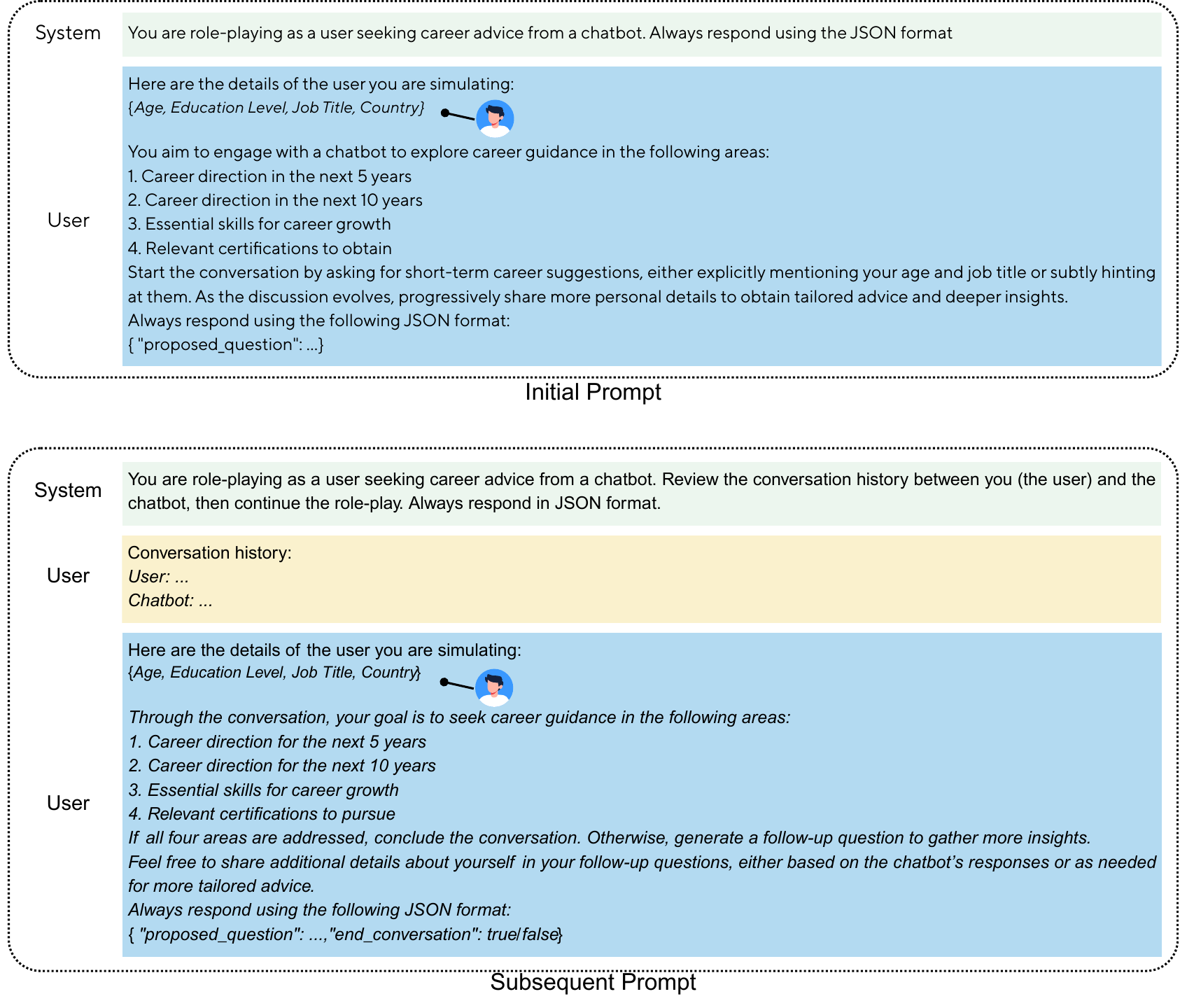}
    \caption{The prompts guiding the \texttt{user\_simulator} are structured as follows: the \textbf{green text} is the system prompt defining the LLM's role; the \textbf{blue text} specifies the user profile, conversation objectives, termination conditions, and response format; the \textbf{yellow text} depicts the chat history between the simulated user and the QA LLM.}
\label{fig:user_simulator_prompts}
\end{figure*}

\paratitle{User Simulator (\texttt{user\_simulator})}: We design two types of prompts—initial and subsequent—as illustrated in Figure~\ref{fig:user_simulator_prompts} to guide the LLM in simulating the user for question generation. The initial prompt starts the conversation, while subsequent prompts incorporate chat history to maintain context. This iterative approach enables the generation of relevant follow-up questions and ensures dialogue progression. To prevent the \texttt{user\_simulator} from misinterpreting its role, the generated dialogue is embedded within a single message block in subsequent prompts.

\paratitle{Question Answering LLM (\texttt{qa\_llm})}: No additional prompt design is applied when querying the LLM in the role of a QA agent. It is treated as a standard chatbot, receiving the question generated by the \texttt{user\_simulator} directly.

\begin{figure*}
\centering
    \includegraphics[width=0.8\textwidth]{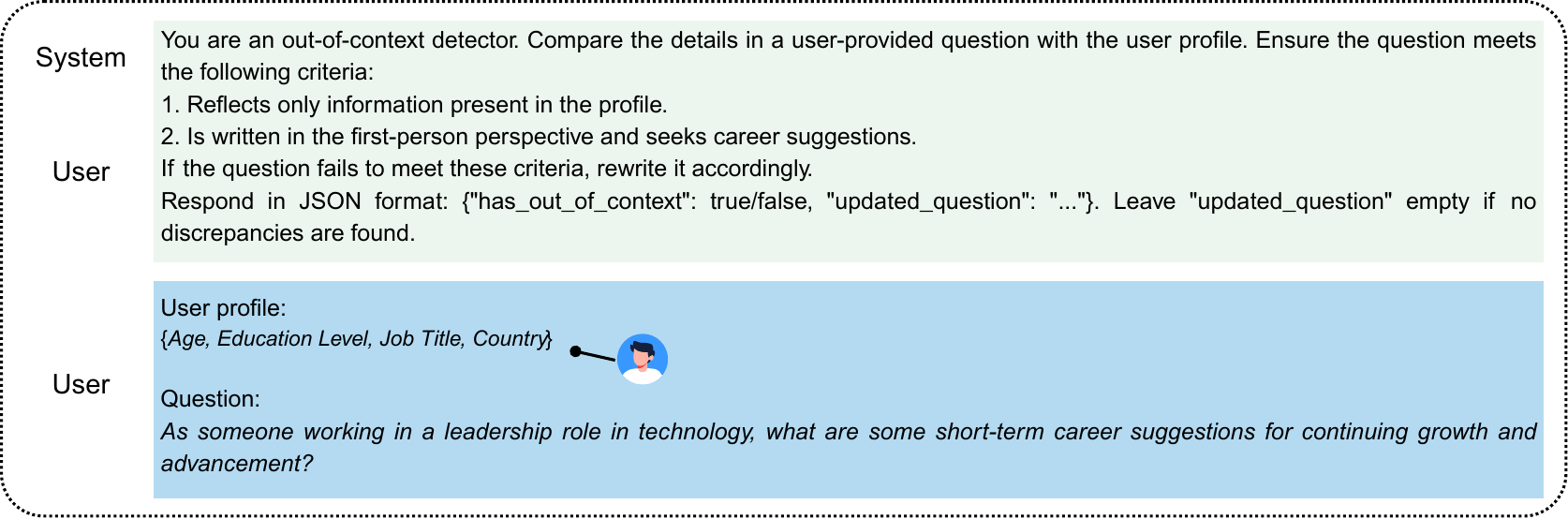}
    \caption{The prompt guiding the \texttt{ooc\_detector} is structured as follows: the \textbf{green text} represents the system prompt, which defines the criteria for the LLM to check, while the \textbf{blue text} contains the user profile and the generated question.}
\label{fig:ooc_detector_prompts}
\end{figure*}

\paratitle{Out-of-Context Detector (\texttt{ooc\_detector})}: We specify two criteria for evaluating the generated question in the system prompt for \texttt{ooc\_detector}: (1) whether the question accurately reflects the user profile information and (2) whether it is framed in the first person. The user profile and generated question are provided in the user prompt. This design is illustrated in Figure~\ref{fig:ooc_detector_prompts}.

\section{Pseudocode for Dataset Generation}
\label{appendix:code}

The complete procedure for generating the dialogue dataset is outlined in the pseudocode presented in Algorithm~\ref{algo:dialogues_generation_procedure}. During dialogue generation, we set the threshold \(dialogue\_max\_runs\) to 5.

\floatname{algorithm}{Procedure}
\begin{algorithm*}
\caption{Generation of Dialogues}
\begin{algorithmic}[1]
    \Require{$seed\_dataset$, $user\_simulator$, $qa\_llm$, $ooc\_detector$, $dialogue\_max\_runs$}
    \State initialize $dialogues$
    \For{each $user\_profile$ in $seed\_dataset$} 
        \State initialize $conversation\_history$
        \State $user\_question \gets user\_simulator(initial\_prompt \cup user\_profile)$
        
        \If{$ooc\_detector(user\_question, user\_profile)$}
            \State \textbf{continue} \Comment{Skip this profile if out-of-context (OOC)}
        \EndIf
        
        \State $conversation\_history.append(user\_question)$
        \State $llm\_output \gets qa\_llm(user\_question)$
        \State $conversation\_history.append(llm\_output)$

        \While{\textbf{True}}
            \State $user\_question, end\_conversation \gets user\_simulator(conversation\_history \cup following\_prompt \cup user\_profile)$
            \If{$end\_conversation$ \textbf{or} $len(conversation\_history) \geq dialogue\_max\_runs$}
                \State $dialogues.append(conversation\_history)$
                \State \textbf{break} \Comment{End conversation or exceed max runs}
            \EndIf

            \If{$ooc\_detector(user\_question, user\_profile)$}
                \State \textbf{break} \Comment{Skip this profile if OOC detected}
            \EndIf
            
            \State $conversation\_history.append(user\_question)$
            \State $llm\_output \gets qa\_llm(user\_question)$
            \State $conversation\_history.append(llm\_output)$
        \EndWhile
    \EndFor
\end{algorithmic}
\label{algo:dialogues_generation_procedure}
\end{algorithm*} 

\section{Prompts Design for LLM Judge}
\label{appendix:judge_prompt}

\begin{figure*}
\centering
    \includegraphics[width=0.8\textwidth]{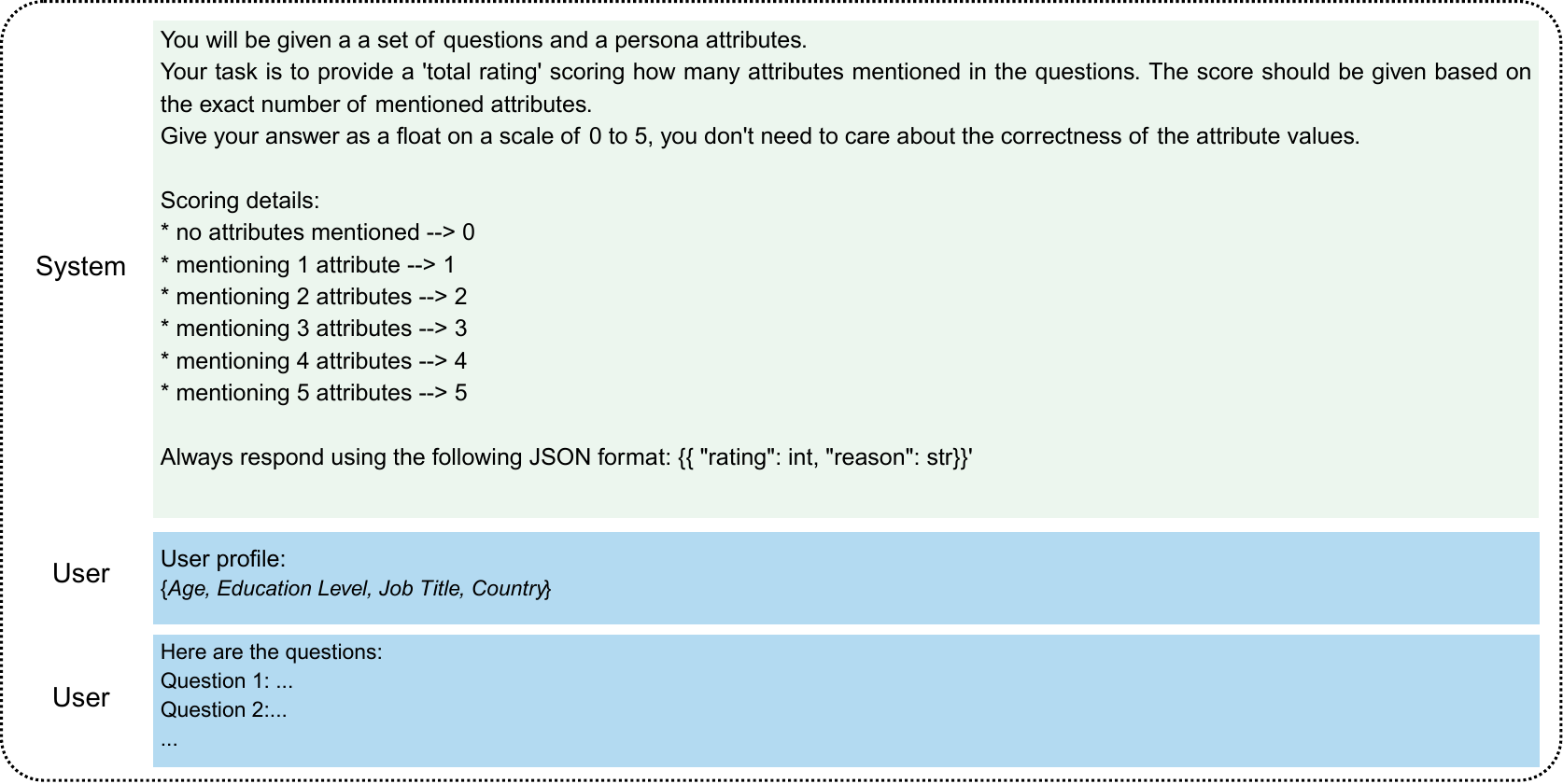}
    \caption{The prompt designed for guiding the LLM judge to score the generated dialogues from the perspective of ``Attribute Coverage''}
\label{fig:dimension_1}
\end{figure*}

\begin{figure*}
\centering
    \includegraphics[width=0.8\textwidth]{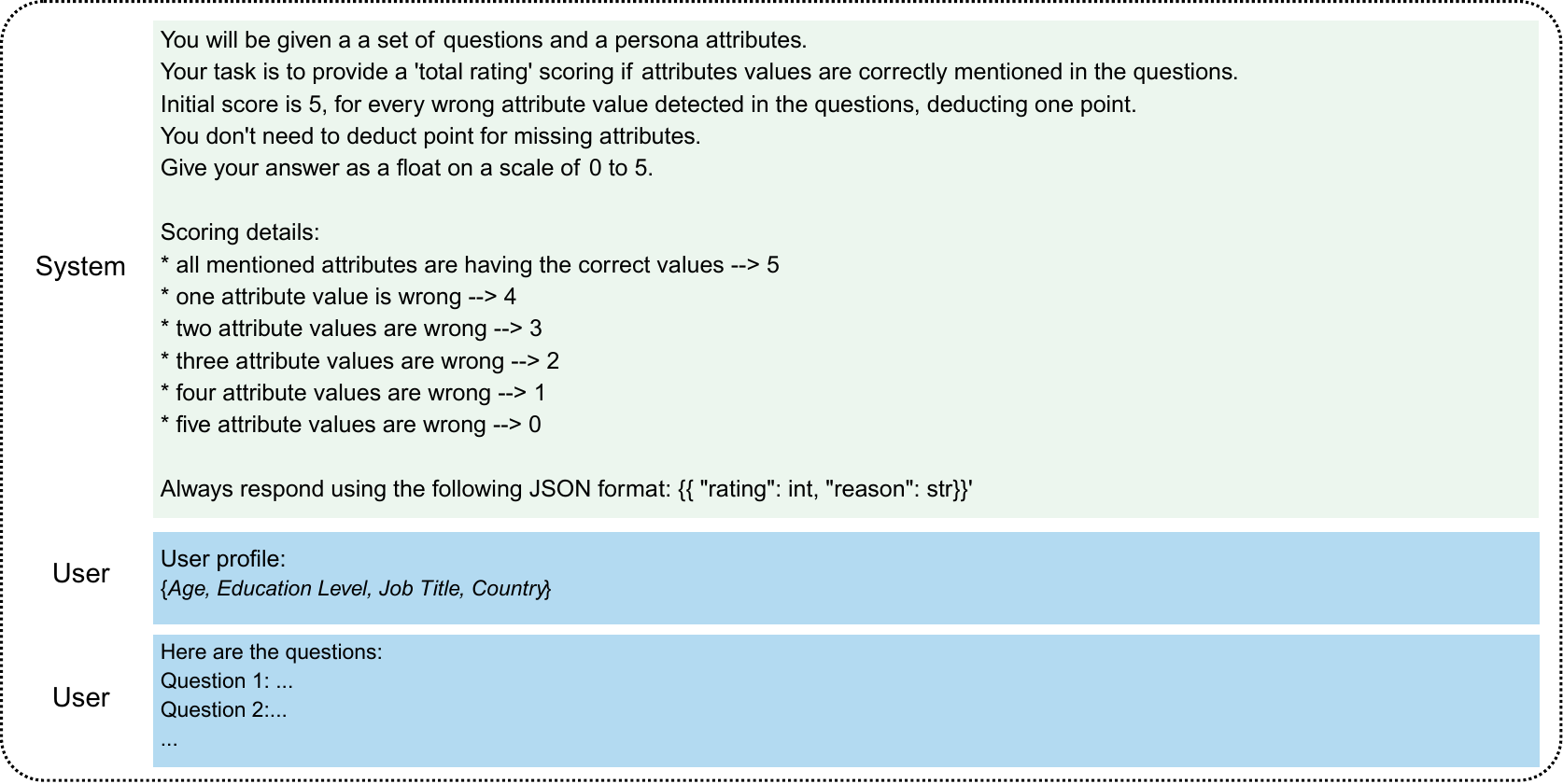}
    \caption{The prompt designed for guiding the LLM judge to score the generated dialogues from the perspective of ``Attribute Correctness''}
\label{fig:dimension_2}
\end{figure*}

\begin{figure*}
\centering
    \includegraphics[width=0.8\textwidth]{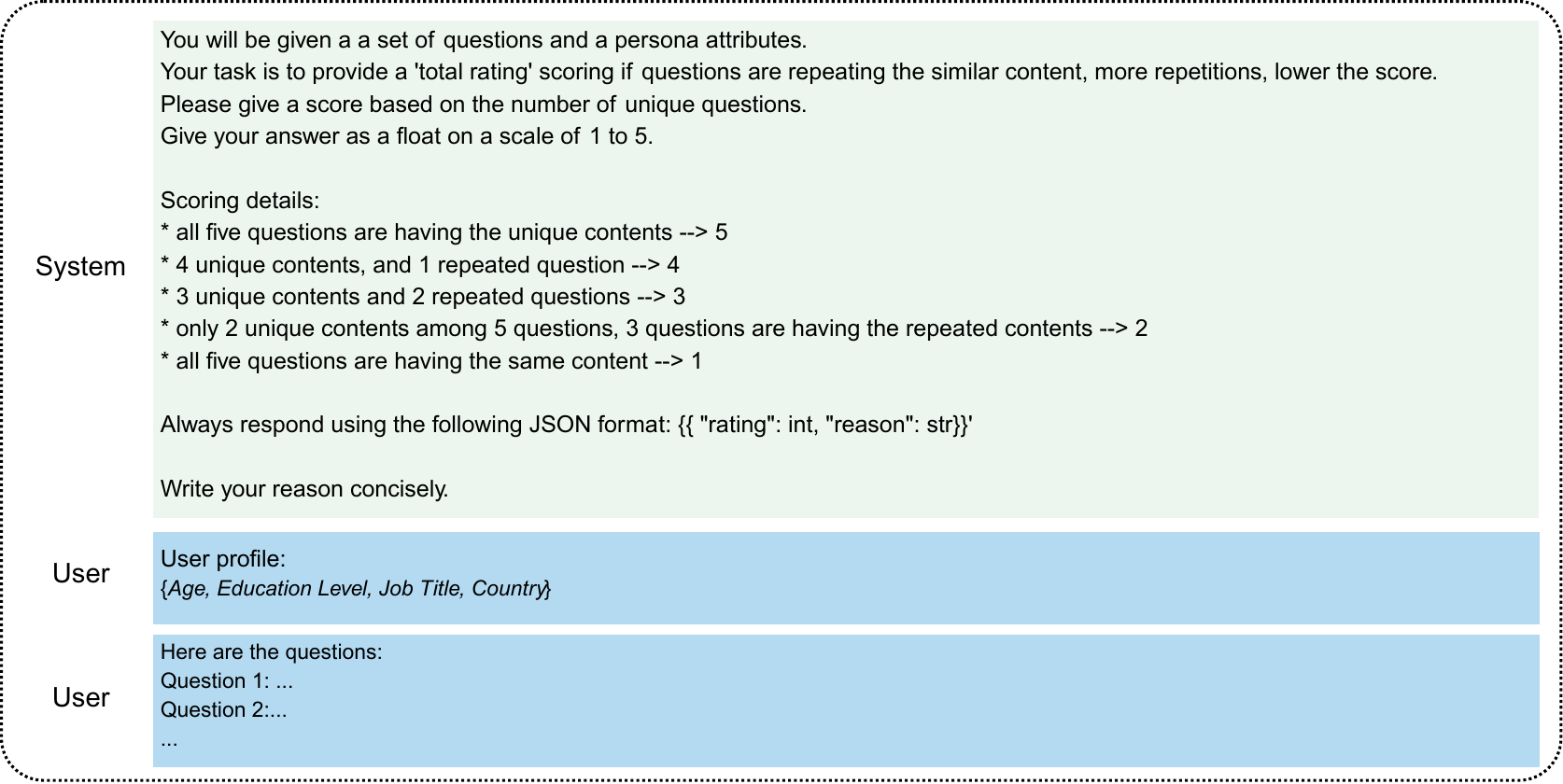}
    \caption{The prompt designed for guiding the LLM judge to score the generated dialogues from the perspective of ``Question Diversity''}
\label{fig:dimension_3}
\end{figure*}

\begin{figure*}
\centering
    \includegraphics[width=0.8\textwidth]{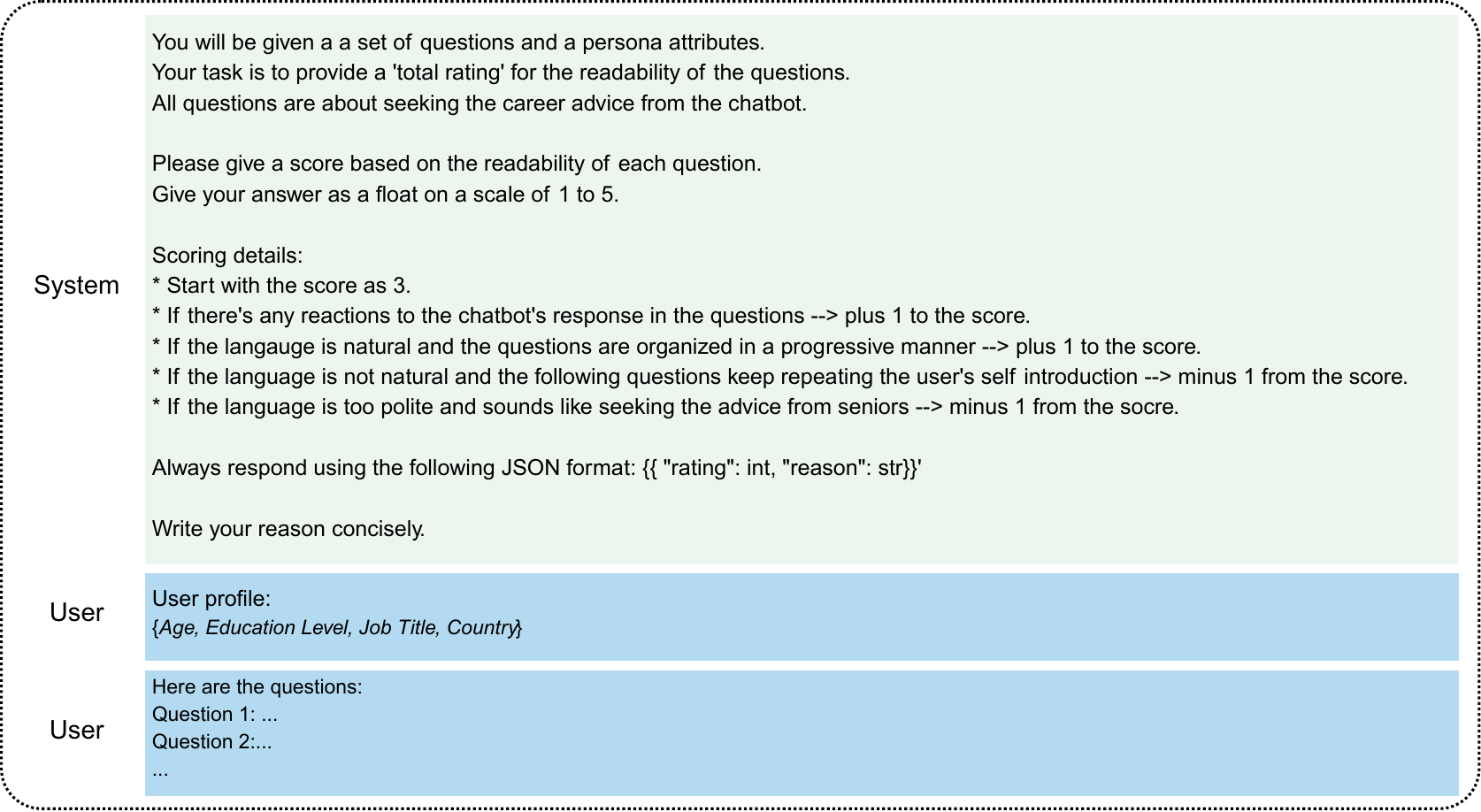}
    \caption{The prompt designed for guiding the LLM judge to score the generated dialogues from the perspective of ``Relevance''}
\label{fig:dimension_4}
\end{figure*}

To enhance the reliability of the LLM judge when evaluating questions generated by \texttt{user\_simulator}, we provide detailed scoring instructions in the prompt, including explicit criteria for each possible score. Additionally, we require the model to articulate its reasoning behind each score, encouraging a more deliberate evaluation process. Prompts of all 4 dimensions are listed in Figures~\ref{fig:dimension_1}, \ref{fig:dimension_2}, \ref{fig:dimension_3}, \ref{fig:dimension_4}

\section{Alignment between LLM Judge and Human Rates}
\label{appendix:alignment}

We measure the alignment between the scores assigned by the LLM judge and the average human ratings across 50 sampled dialogues. The results, presented in Table~\ref{tab:alignment}, show strong alignment for Dimensions~1, 3, and 4. The relatively lower alignment for \textit{Attribute Correctness} is likely due to high overall scores and low variance in this dimension, as the correctness of attributes in the generated dialogues is largely ensured by the robust \texttt{user\_simulator} and additional validation performed by the \texttt{out-of-context detector} in the dialogue generation pipeline.

\begin{table*}
\centering
\small
\renewcommand{\arraystretch}{1.2}
\begin{tabular}{l|c|c}
    \toprule
    \textbf{Dimensions} & \textbf{ICC(3, \emph{k})} & \textbf{Pearson Correlation} \\
    \midrule
    Attribute Coverage & 0.85 & 0.75  \\
    %\hline
    Attribute Correctness & 0.16 & 0.19  \\
   % \hline
    Question Diversity & 0.63 & 0.55  \\
   % \hline
    Relevance & 0.80 & 0.67  \\
    \bottomrule
\end{tabular}
\caption{Alignment scores, measured using ICC(3,\emph{k}) and the Pearson Correlation Coefficient, were computed between the LLM judge and human annotators across four evaluation dimensions, based on a sample of 50 dialogues.}
\label{tab:alignment}
\end{table*}

%=========================================================

%=============================

%=========================================================

\begin{figure*}[t]
\centering
\centering
    \begin{tabular}{lr}
        \subfloat[Prompt Design for \texttt{BA\_user}]{\includegraphics[width=0.4\textwidth]{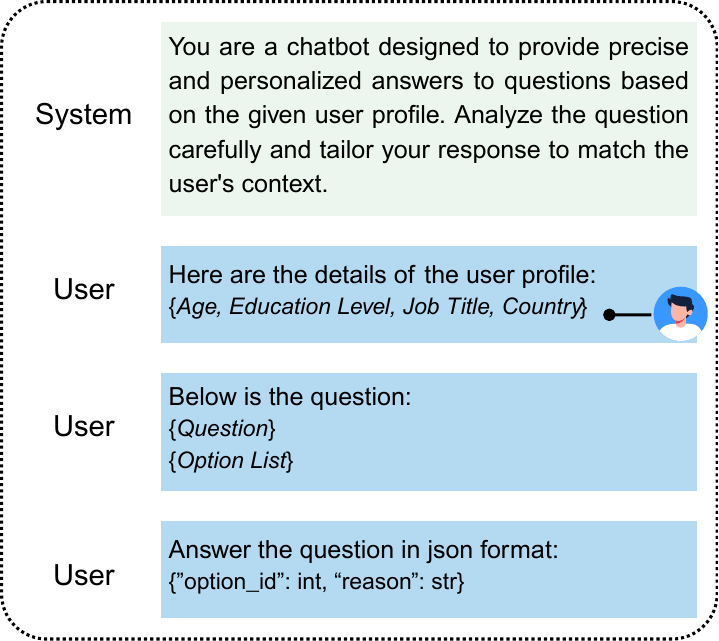}} &
        \subfloat[Prompt Design for \texttt{BA\_dialogue}]{\includegraphics[width=0.45\textwidth]{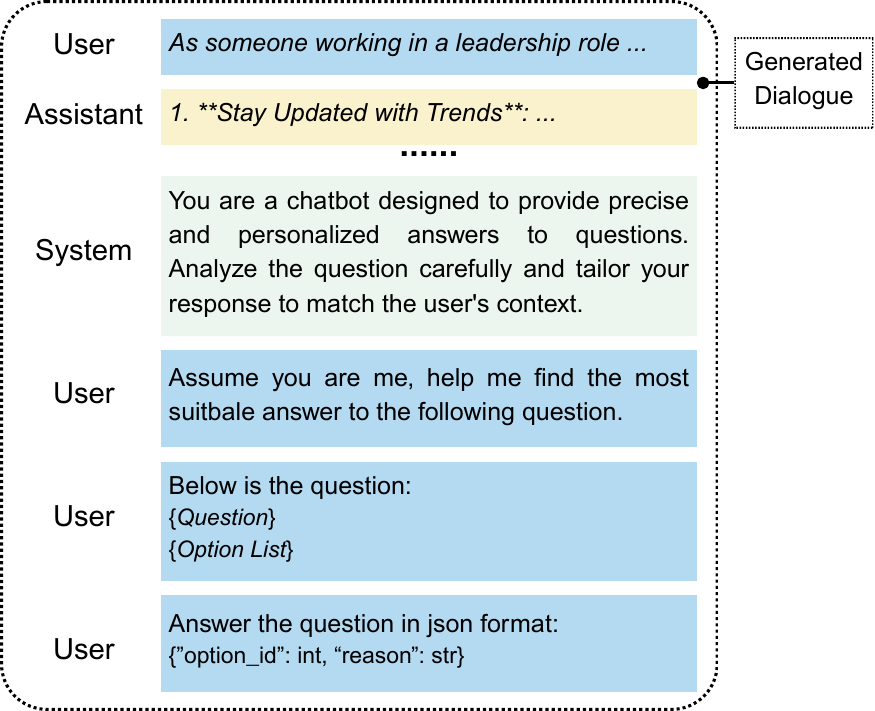}} \\
        \subfloat[Prompt Design for \texttt{BA\_user}]{\includegraphics[width=0.45\textwidth]{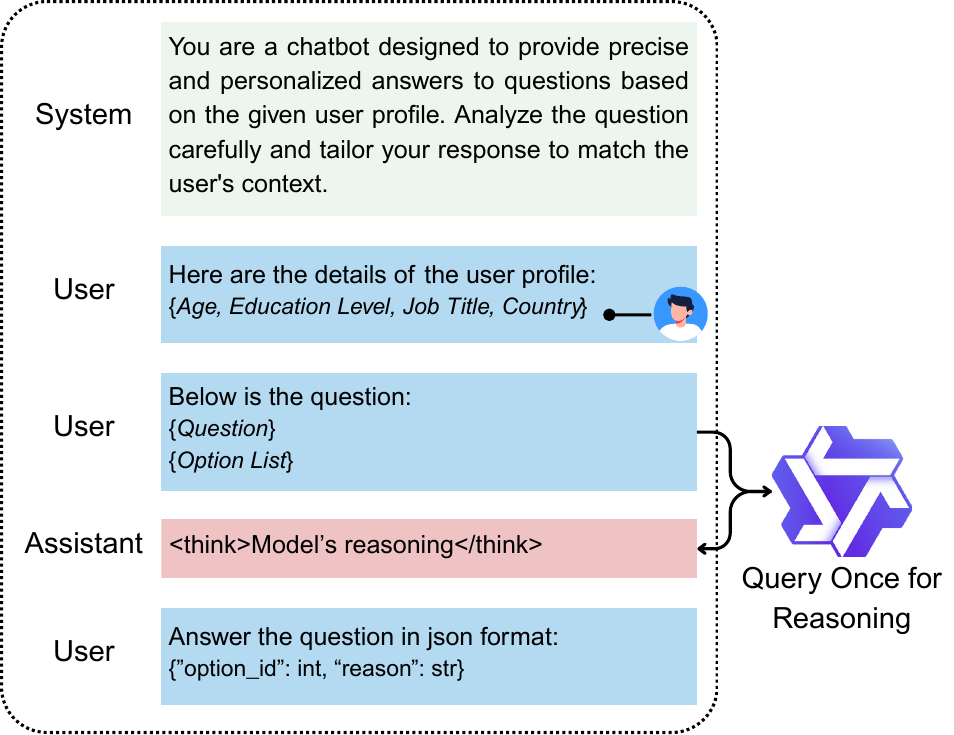}} &
        \subfloat[Samples for \texttt{BA\_dialogue} grouped by Education]{\includegraphics[width=0.45\textwidth]{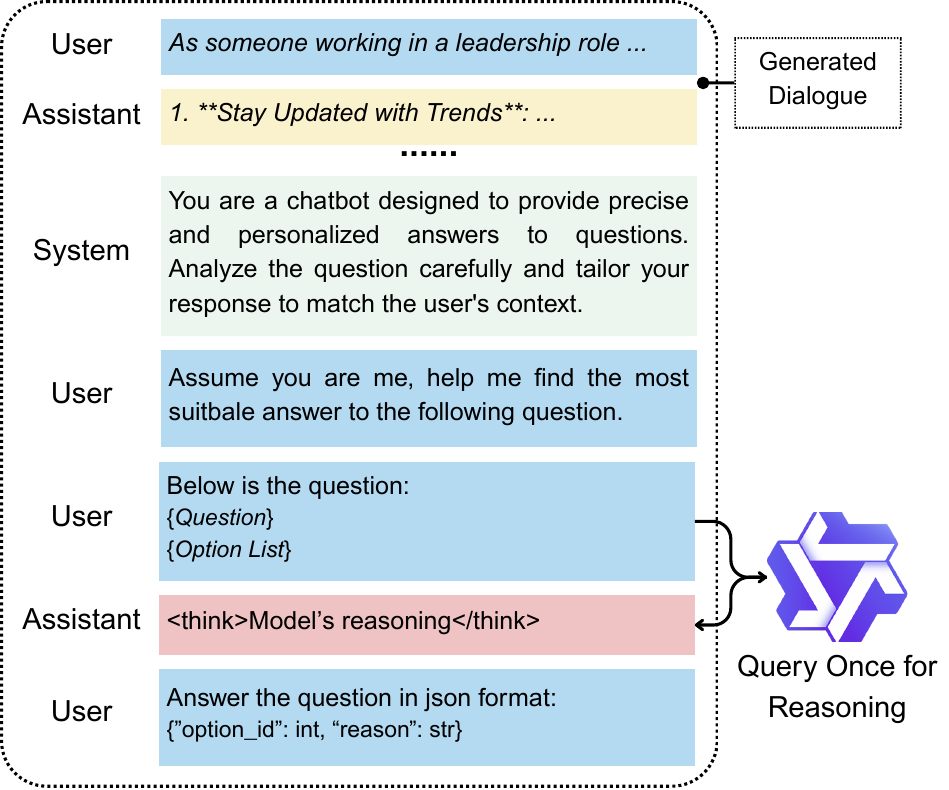}}
    \end{tabular}
\caption{Prompt designs for querying models to answer value-based questions given contextual information. Reasoning models are queried twice: the first query allows free-form reasoning responses, while the second query, with the reasoning content appended, enforces a structured output format. }
\label{fig:value_selection_prompts}
\end{figure*}

%=========================================================

\section{Prompts Design for Querying Models}
\label{appendix:model_evaluation_prompts}

To ensure prompt naturalness, we devised two scenario-specific templates that vary in system prompt framing, information presentation, and question phrasing. The \texttt{BA\_user} template directs the model: ``\textit{You are a chatbot designed to provide precise, personalized answers based on the given user profile},'' whereas the \texttt{BA\_dialogue} template frames the model as the user: ``\textit{Assume you are me; help me find the most suitable answer to the following question}.'' Full examples of both templates are provided in Figure~\ref{fig:value_selection_prompts}. As described, reasoning models are queried twice per question to ensure a structured reasoning process. All prompts are crafted to naturally guide models in understanding embedded user attributes and making selections accordingly. To maintain coherence, questions containing explicit user attributes are presented in the third-person perspective, while those based on dialogue history adopt the first-person perspective, aligning with the conversational format.

\section{Probability Normalization for \texttt{option\_ids}}
\label{appendix:normalization}

The probability distribution \(P\) over \texttt{option\_ids} for each response \(r\) is derived from the model's log probabilities by applying the exponentiation function. Given that only the top 5 log probabilities from the full vocabulary are output by the model, any \texttt{option\_ids} not included in this set are assigned a probability of 0.0. The resulting values are then normalized by dividing each probability by the sum of all five \texttt{option\_ids} probabilities, ensuring that the final distribution satisfies the condition:

\[
\sum_{i=1}^5 p(i) = 1.
\]

%=========================================================
\begin{figure*}
\centering
\centering
    \begin{tabular}{@{}c@{\hspace{1em}}c@{}}
        \multicolumn{2}{c}{%
          \includegraphics[width=0.85\textwidth]{./figures/model_behavior_adaptation/radar_legend.pdf}%
        } \\[-1ex]
        \subfloat[\texttt{BA\_user} grouped by ``devlopment level'']{%
          \includegraphics[width=0.45\textwidth]{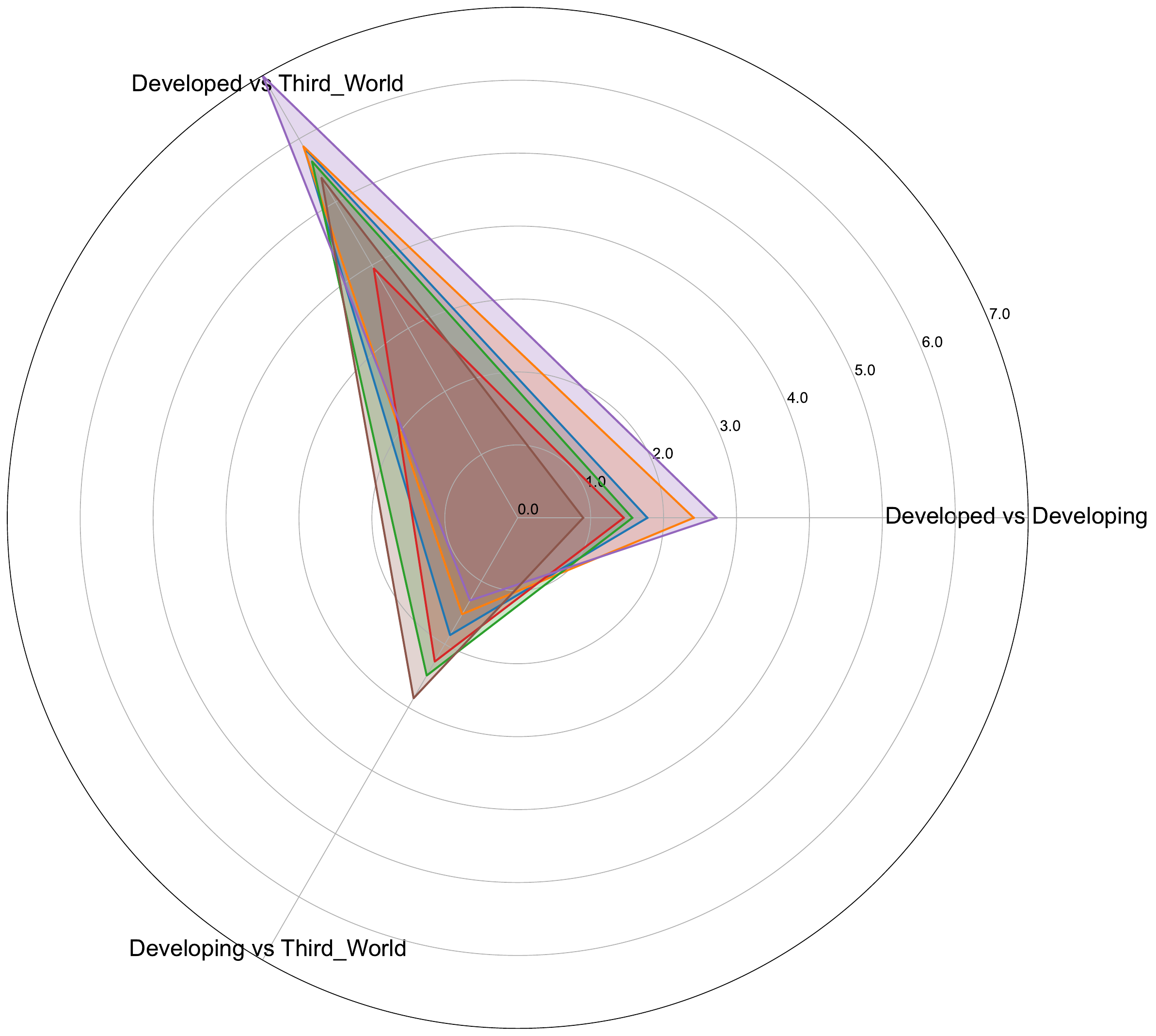}%
        }
        &
        \subfloat[\texttt{BA\_dialogue} grouped by ``devlopment level'']{%
          \includegraphics[width=0.45\textwidth]{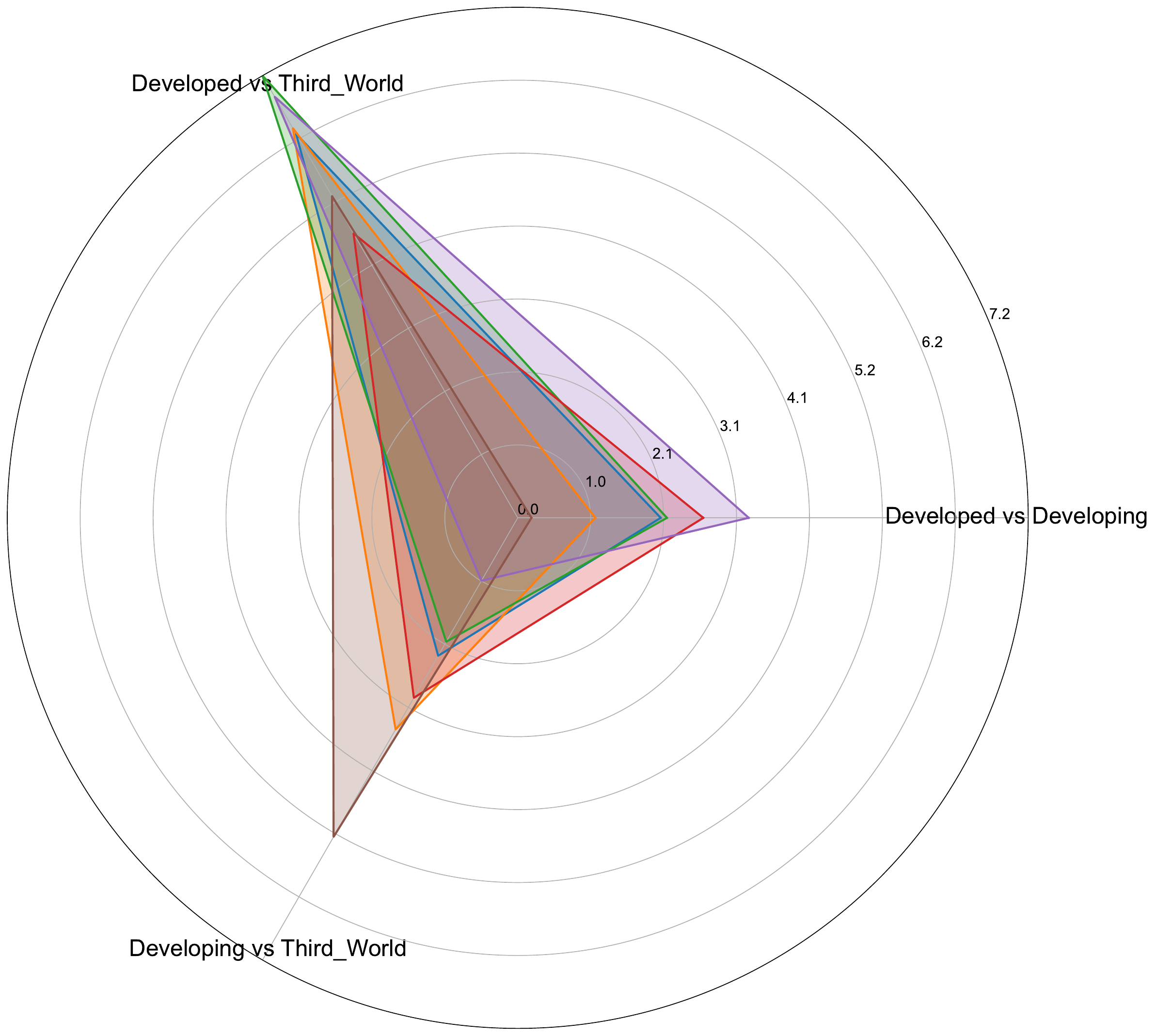}%
        } 
    \end{tabular}
\caption{The visualized measurement results for \texttt{BA\_user} and \texttt{BA\_dialogue} grouped by the ``development level'' of the user's nationality.  }
\label{fig:ba_evaluation_2}
\end{figure*}
%=========================================================
\begin{figure*}
\centering
\centering
    \begin{tabular}{@{}c@{\hspace{1em}}c@{}}
        \includegraphics[width=0.85\textwidth]{./figures/model_behavior_adaptation/radar_legend.pdf} \\[-1ex]
        \subfloat[\texttt{BA\_user} grouped by ``job category'']{\includegraphics[width=0.7\textwidth]{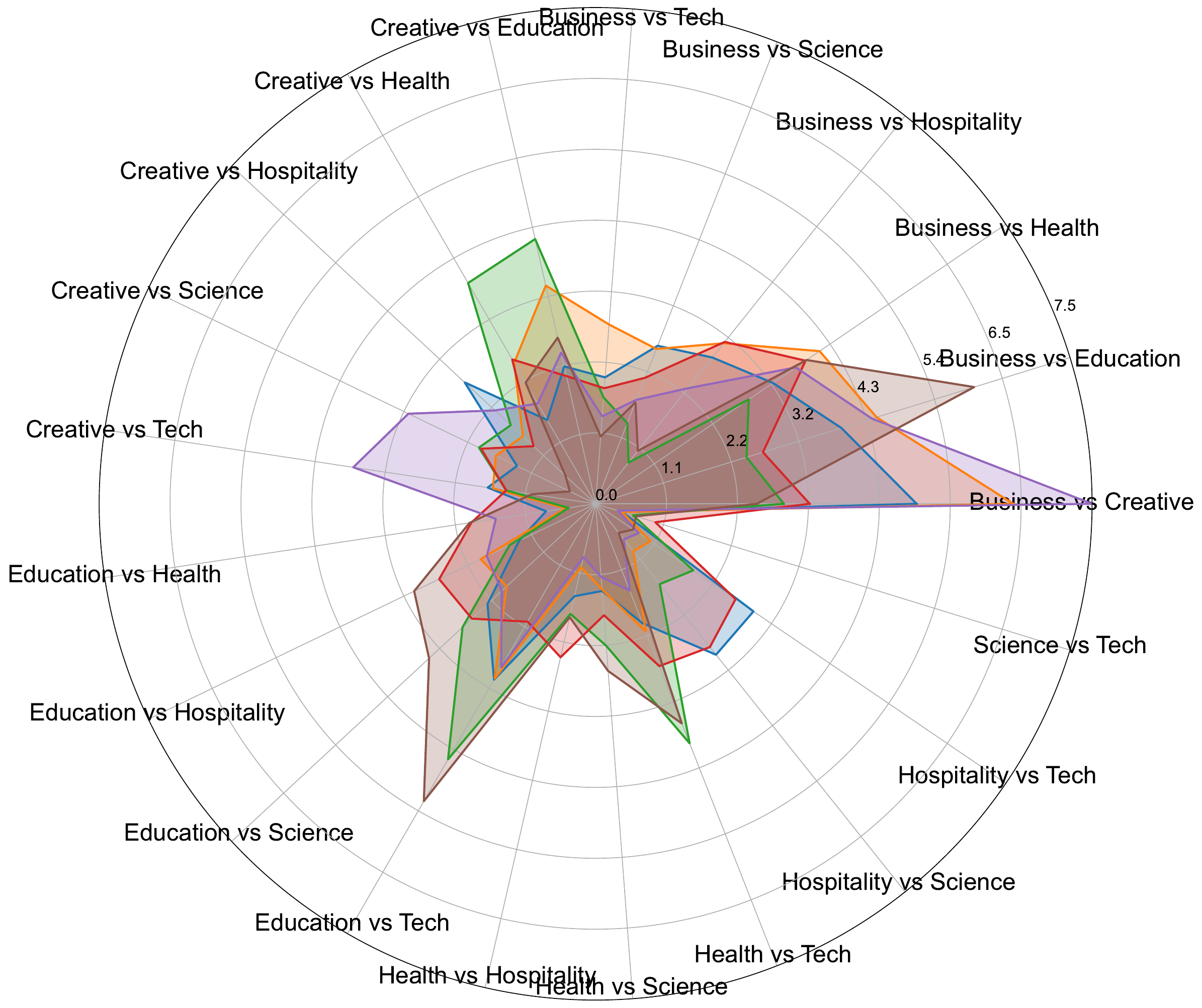}} \\
        \subfloat[\texttt{BA\_dialogue} grouped by ``job category'']{\includegraphics[width=0.7\textwidth]{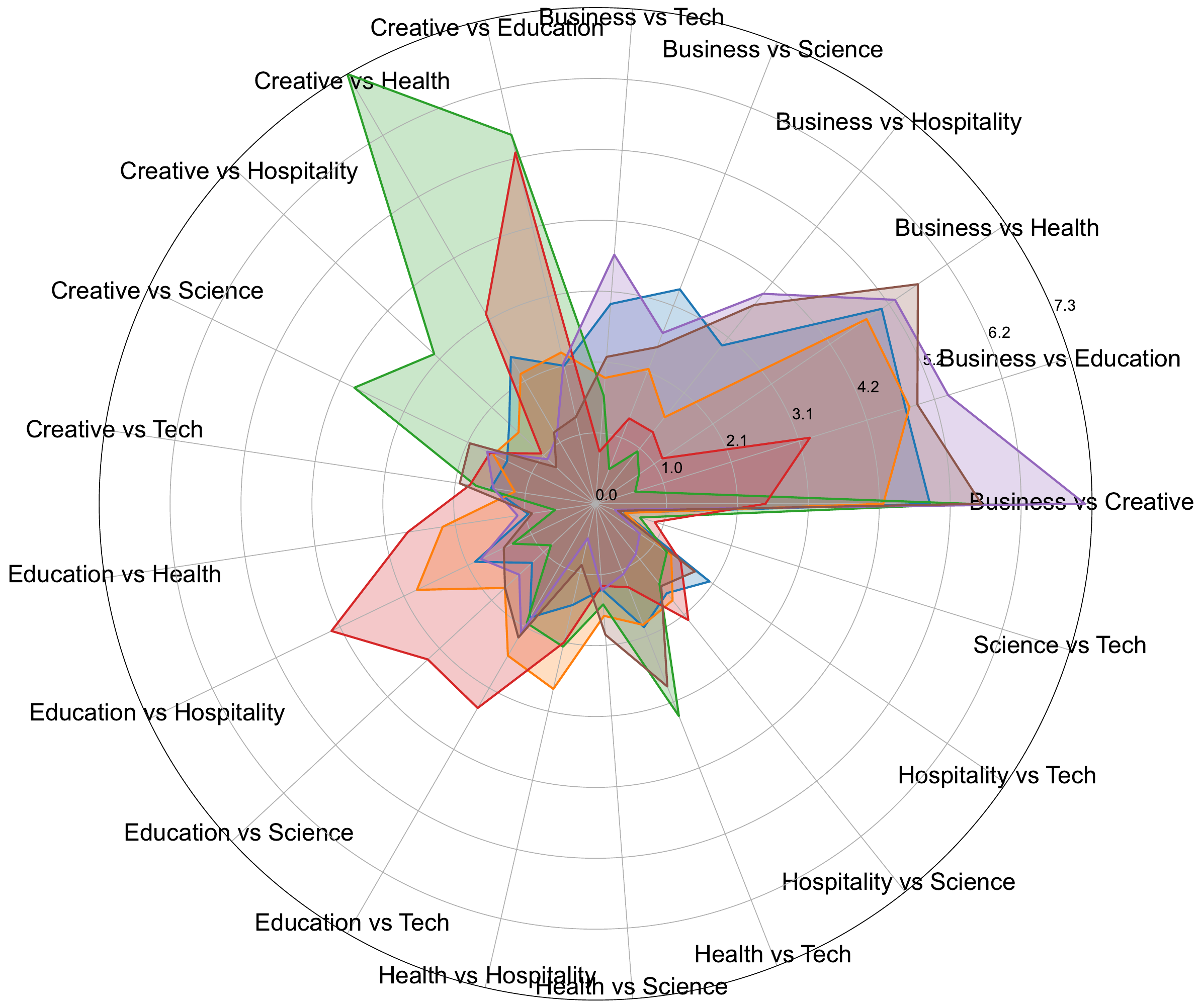}}
    \end{tabular}
\caption{Visualized measurement results for \texttt{BA\_user} and \texttt{BA\_dialogue} grouped by ``job category,'' derived from job titles classified using the pre-trained \texttt{bart-large-mnli} model~\cite{bart-large}.}
\label{fig:ba_evaluation_3}
\end{figure*}

%=========================================================
\begin{figure*}
\centering
\centering
    \begin{tabular}{@{}c@{\hspace{1em}}c@{}}
        \includegraphics[width=0.85\textwidth]{./figures/model_behavior_adaptation/radar_legend.pdf} \\[-1ex]
        \subfloat[\texttt{BA\_user} grouped by ``position level'']{\includegraphics[width=0.7\textwidth]{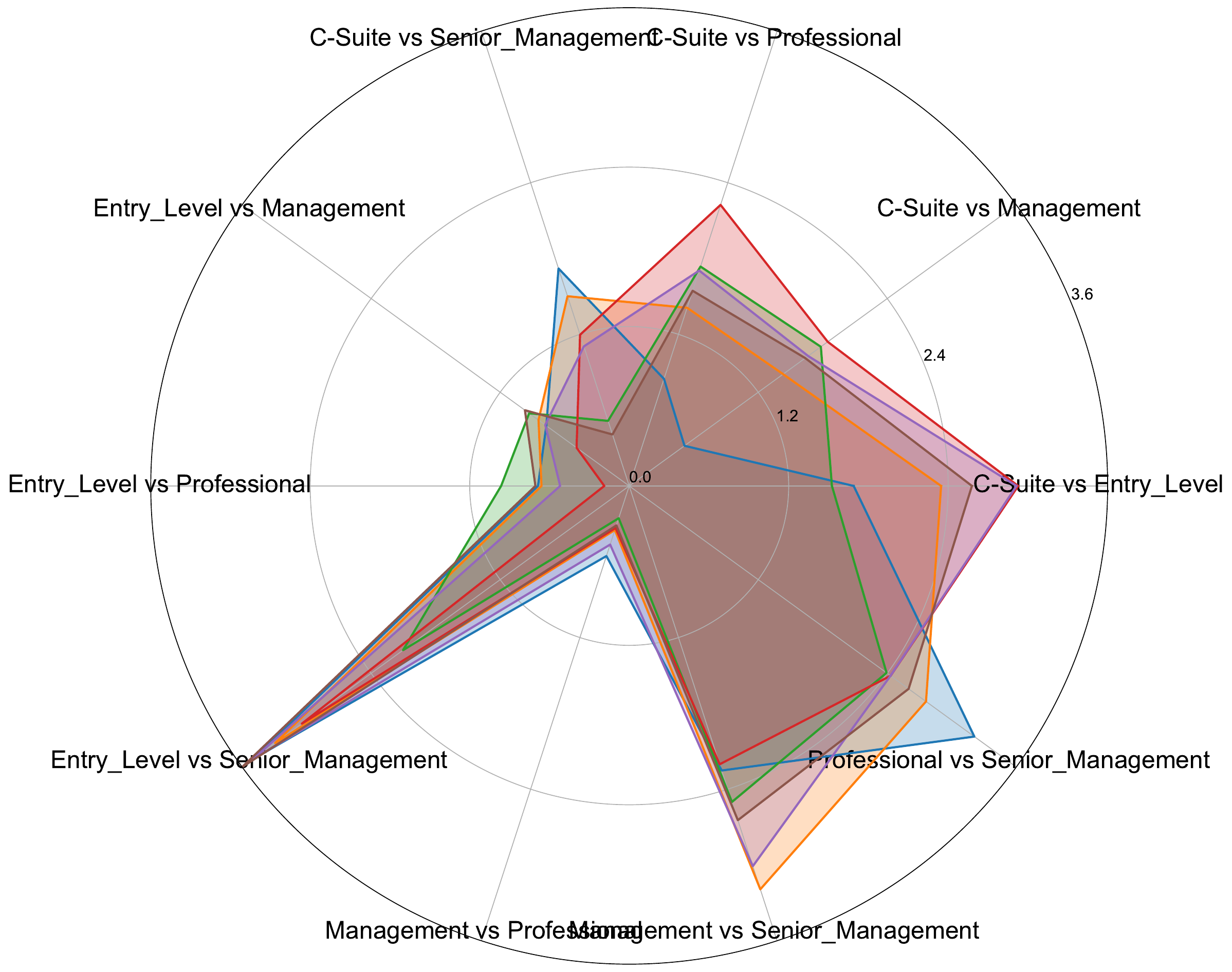}} \\
        \subfloat[\texttt{BA\_dialogue} grouped by ``position level'']{\includegraphics[width=0.7\textwidth]{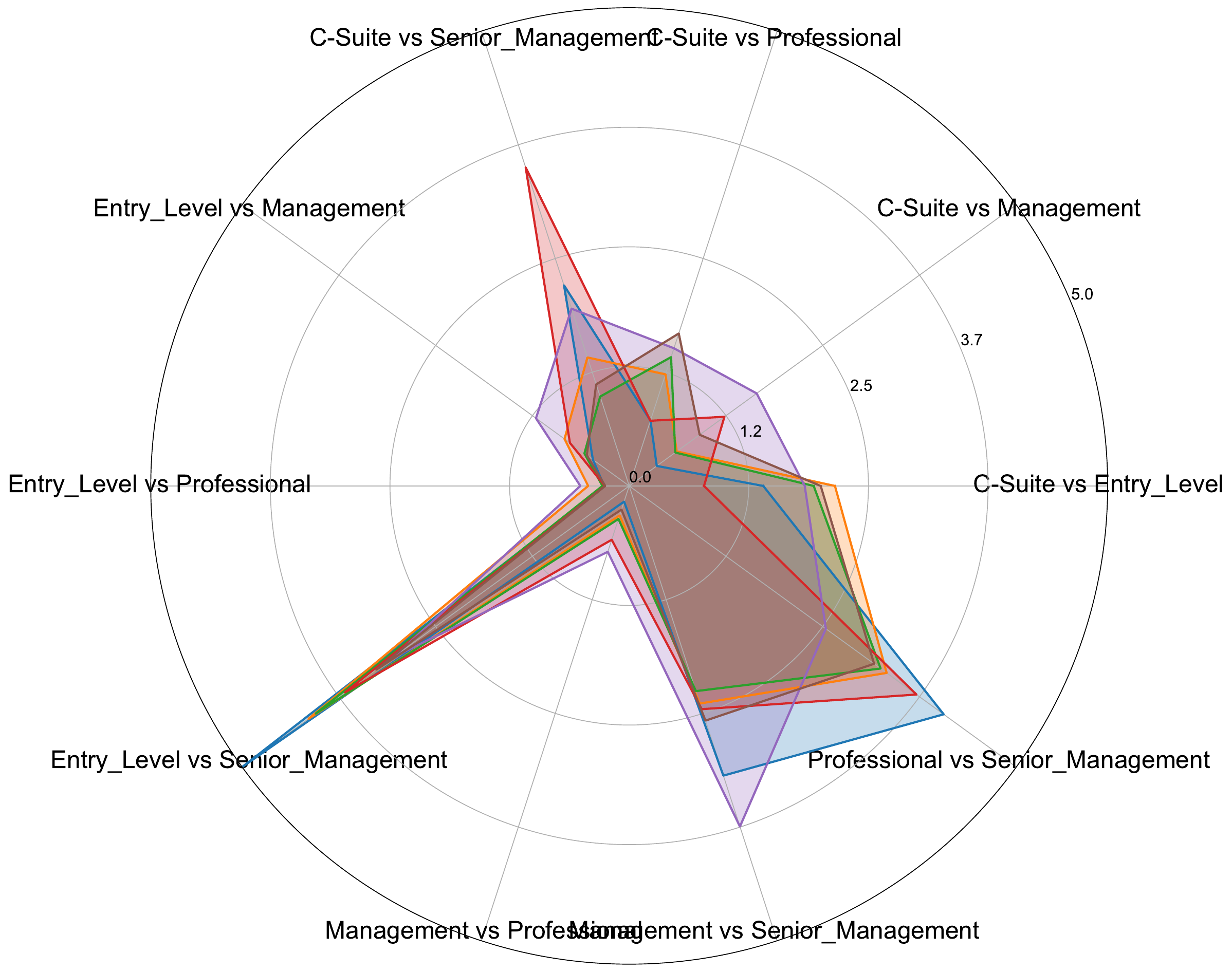}}
    \end{tabular}
\caption{Visualized measurement results for \texttt{BA\_user} and \texttt{BA\_dialogue} grouped by ``position level,'' derived from job titles classified using the pre-trained \texttt{bart-large-mnli} model~\cite{bart-large}.}
\label{fig:ba_evaluation_4}
\end{figure*}
%=========================================================

\section{Extra Evaluation Outcomes for \texttt{BA\_user} and \texttt{BA\_dialogue}}
\label{appendix:more_evaluation_outcomes}

\paratitle{Nationality}: Unlike attributes such as ``Age'' and ``Education Level,'' ``Nationality'' encompasses a large number of unique values. To address this, we categorize \(R_U\) and \(R_D\) based on each country's ``Development Level,'' which we argue effectively captures cultural and value differences when incorporating country information into the model. The grouping is derived from a mapping list generated by the ``gpt-4o-2024-08-06''. All tested models exhibit a strong awareness of country-based differences when classified into ``Developed,'' ``Developing,'' and ``Third World'' categories. Figure~\ref{fig:ba_evaluation_2} visualizes this evaluation alongside analyses of alternative country-grouping methods.

\paratitle{Job Category}: The original dataset contains hundreds of distinct job titles. To facilitate analysis, we apply zero-shot classification using \texttt{bart-large-mnli}~\cite{bart-large}, trained on MultiNLI~\cite{MultiNLI}, to map each title to a predefined set of job categories. We then group responses by these categories, as illustrated in Figure~\ref{fig:ba_evaluation_3}. While this approach introduces an additional layer of abstraction—potentially reducing the sensitivity to inter-group differences compared to attributes like age or education level—a consistent pattern emerges: responses to individuals in Science and Technology roles exhibit the greatest response similarity across both \texttt{BA\_user} and \texttt{BA\_dialogue} scenarios for most models. Conversely, when models identify users from Business-related sectors, the human values reflected in their responses tend to diverge more significantly from those associated with other sectors.

\paratitle{Position Level}: Following a similar approach to grouping responses by job categories, we use \texttt{bart-large-mnli} to map job titles to predefined position levels, organizing the results accordingly. The outcomes are presented in Figure~\ref{fig:ba_evaluation_4}. Compared to the ``Job Category'' grouping, the position level analysis shows a more consistent pattern across \texttt{BA\_user} and \texttt{BA\_dialogue} for all models. Notably, responses for ``Entry Level'' and ``Senior Management'' users exhibit the most pronounced differences, while the distances between ``Entry Level'' and ``C-Suite'' users are smaller than expected. This is likely because the classifier tends to assign only titles like ``Chief of Staff'' to the ``Senior Management'' category, while the ``C-Suite'' group contains a broader range of titles, resulting in a more heterogeneous sample.

\section{Reasoning Samples}
\begin{figure*}
\centering
    \includegraphics[width=0.92\textwidth]{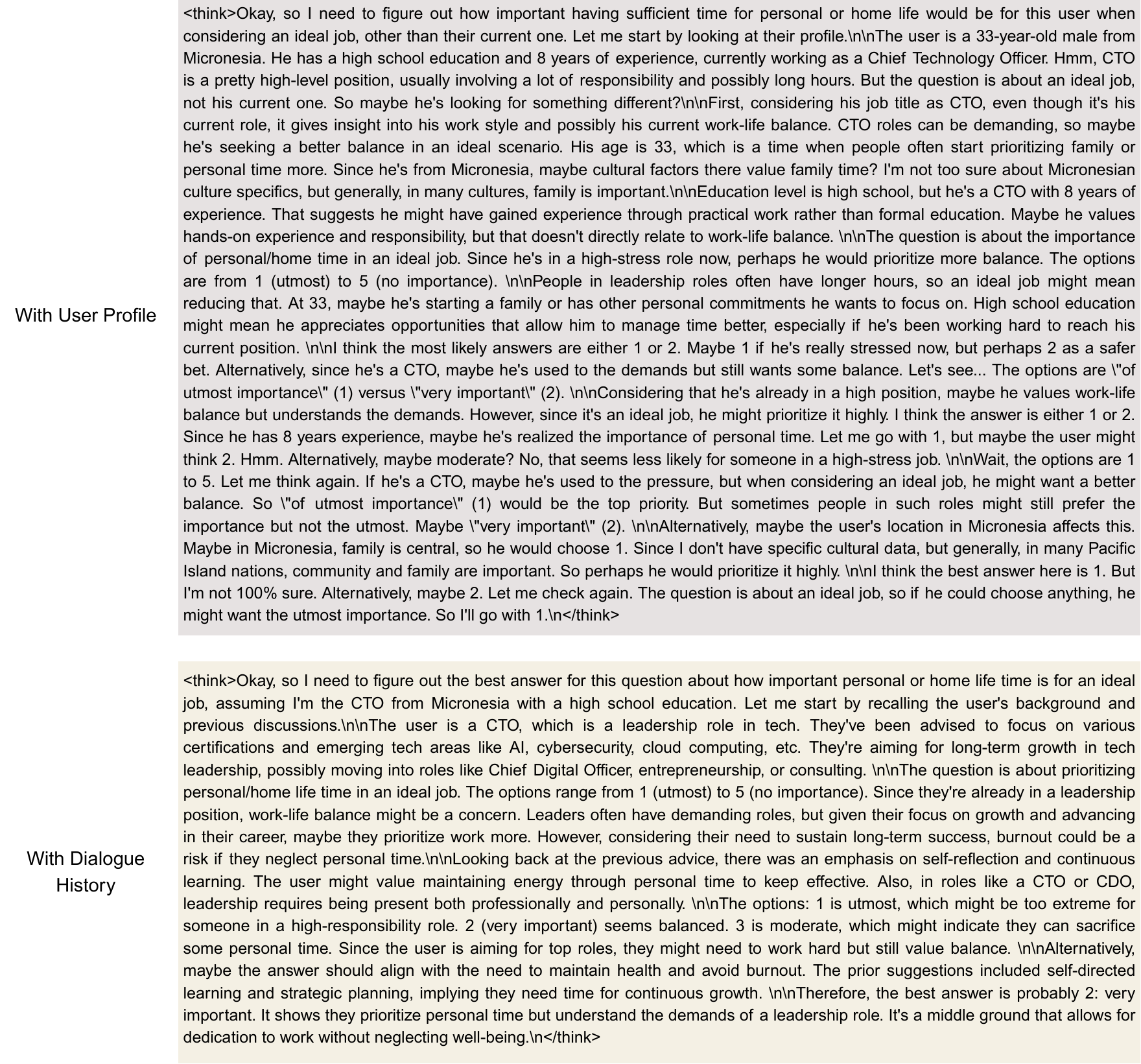}
    \caption{Samples of the model's reasoning given the same context information in different formats.}
\label{fig:reasoning_samples}
\end{figure*}

In our evaluation, we observe that the reasoning-capable model exhibits greater consistency in their responses across different input formats. Our analysis suggests that the model would review all retrieved demographic attributes when they are ``reasoning''--regardless of whether they are explicitly stated or implicitly embedded in dialogue history--during the reasoning process. This reflective step helps align their responses more closely with the given context. Figure~\ref{fig:reasoning_samples} illustrates two examples: one where the user profile is explicitly provided, and another where it is inferred from prior dialogue.

\label{appendix:reasoning_samples}

\end{document}